\documentclass[journal]{IEEEtran}
\ifCLASSINFOpdf
\else
\fi

\usepackage{epsfig}
\usepackage{graphicx}
\usepackage{amsmath}
\usepackage{amssymb}

\usepackage{algorithm}
\usepackage{algorithmicx}
\usepackage{algpseudocode}
\usepackage{mathrsfs}
\usepackage{multirow}
\usepackage{amsthm}

\usepackage{bm}
\usepackage{bbold}
\usepackage{subcaption}
\usepackage{array}

\newtheorem{theorem}{Theorem}
\newtheorem{lemma}{Lemma}
\newtheorem{corollary}{Corollary}

\newtheorem{remark}{Remark}

\usepackage{scalerel}

\usepackage{threeparttable}
\usepackage{xcolor}

\usepackage{makecell}

\begin{document}

\title{Robust Non-Linear Matrix Factorization for Dictionary Learning, Denoising, and Clustering}


\author{Jicong Fan, Chengrun Yang, Madeleine Udell
\thanks{Jicong Fan is with the School of Data Science, The Chinese University of Hong Kong (Shenzhen) and Shenzhen Research Institute of Big Data, Shenzhen, China.
Email: fanjicong@cuhk.edu.cn

Chengrun Yang and Madeleine Udell are with the School of Operations Research and Information Engineering, Cornell University,
Ithaca, NY 14853, USA. Email: $\left\{\textup{cy438,udell}\right\}$@cornell.edu}}


\maketitle
\begin{abstract}
Low dimensional nonlinear structure abounds in datasets across computer vision and machine learning.
Kernelized matrix factorization techniques have recently been proposed to
learn these nonlinear structures for denoising, classification, dictionary learning, and missing data imputation, by observing that the image of the matrix in a sufficiently large feature space is low-rank. However, these nonlinear methods fail in the presence of sparse noise or outliers.
In this work, we propose a new robust nonlinear factorization method
called Robust Non-Linear Matrix Factorization (RNLMF).
RNLMF constructs a dictionary for the data space by
factoring a kernelized feature space;
a noisy matrix can then be decomposed as the sum of a sparse noise matrix
and a clean data matrix that lies in a low dimensional nonlinear manifold.
RNLMF is robust to sparse noise and outliers and scales to matrices with thousands of rows and columns.
Empirically, RNLMF achieves noticeable improvements over baseline methods in denoising and clustering.
\end{abstract}


\section{Introduction}
\IEEEPARstart{R}{eal} data or signals are often corrputed by noise or outliers.
As such, data denoising is a core task across applications in
computer vision, machine learning, data mining and signal processing.
Many denoising strategies exploit the difference between the
distribution of the data (structured, coherent) and the distribution of the noise (independent).
Perhaps the simplest latent structure is low-rank structure,
which appears throughout a wide range of applications \cite{udell2019big}
and undergirds celebrated algorithms such as principal component analysis (PCA) \cite{wold1987principal,PCA_Jolliffe2002},
robust PCA (RPCA) \cite{RPCA}, and low-rank matrix completion \cite{CandesRecht2009,Fan2017290,fan2019factor}.
For instance, RPCA aims to decompose a partially corrupted matrix
as the sum of a low-rank matrix and a sparse matrix, thus separating the sparse noise from the low rank data.
Another well-known latent structure is sparsity, or more specifically,
the property that data vectors can be modeled as a sparse linear combination of basis elements.
Sparse structure appears in compressed sensing \cite{cs_Donoho}, subspace clustering \cite{SSC_PAMIN_2013,LRR_PAMI_2013,FAN201839,you2018scalable}, dictionary learning \cite{mairal2009online,mairal2009supervised,zhang2019joint,zhang2019scalable}, image classification \cite{4483511,zhang2011sparse}, noise/outlier identification \cite{lu2013online,jiang2015robust}, and semi-supervised learning \cite{li2016weakly,li2017robust}. For instance, the self-expressive models widely used in subspace clustering \cite{SSC_PAMIN_2013,LRR_PAMI_2013,LSSLRSC} exploit the fact that each data point can be represented as a linear combination of a few data points lying in the same subspace and hence are able to recognize noise and outliers when the data are drawn from a union of low-dimensional subspaces. In \cite{li2017robust}, the authors proposed to explicitly pursue structured (block-diagonal) sparsity  for robust representation with partially labeled data. In recent years, a number of kernel methods \cite{Mika99kernelpca,kwok2004pre,huang2009robust,nguyen2009robust,fan2019exactly} and deep learning methods \cite{vincent2010stacked,burger2012image,mao2016image,gao2017demand} have been
proposed to remove noise from data with nonlinear low-dimensional latent structures. Nevertheless, the kernel denoising methods have high time and space complexities. Most of the deep-learning-based denoising methods require clean data samples (e.g. noiseless images), as supervision, to train the neural networks. In this paper, we focus on unsupervised denoising. More recently, a few kernelized factorization methods \cite{van2013design,7299018,liu2015robust,quan2016equiangular} have been proposed for nonlinear dictionary learning but they are not able to handle sparse noise. To solve the problem, we in this paper provide a robust nonlinear matrix factorization method and apply it to dictionary learning, denoising, and clustering.

\section{Related work and our contribution}
\subsection{Robust principal component analysis}
Suppose a data matrix $\bm{X}\in\mathbb{R}^{m\times n}$ is partially corrupted
by sparse noise $\bm{E}\in\mathbb{R}^{m\times n}$ to form noisy observations
$\hat{\bm{X}}=\bm{X}+\bm{E}$.
The robust PCA (RPCA) model in \cite{RPCA} assumes that $\bm{X}$ is low-rank and aims to solve
\begin{equation}\label{Eq.RPCA}
\mathop{\textup{minimize}}_{\bm{X},\bm{E}} \ \Vert\bm{X}\Vert_\ast+\lambda\Vert\bm{E}\Vert_1,\quad
\textup{subject to}\ \bm{X}+\bm{E}=\hat{\bm{X}},
\end{equation}
where $\Vert\bm{X}\Vert_\ast$ denotes the matrix nuclear norm
(sum of singular values; a convex relaxation of matrix rank)
of $\bm{X}$.
RPCA is not effective in denoising data with nonlinear latent structure,
as the corresponding matrix $\bm{X}$ is often of high rank. In \cite{fan2019exactly}, a robust kernel PCA (RKPCA) was proposed to remove sparse noise from nonlinear data. The space and time complexities of the naive algorithm are $O(n^2)$ and $O(n^3)$ respectively, to store an $n\times n$ kernel matrix and compute its singular value decomposition (SVD). Thus RKPCA are not scalable to large-scale data. In addition, RKPCA has no out-of-sample extension to reduce the noise of new data efficiently.

\subsection{Robust dictionary learning and subspace clustering}
Classical dictionary learning \cite{aharon2006k} and sparse coding algorithms
\cite{mairal2009online,skretting2010recursive,mairal2011task,6516503}
denoise by projecting onto $d$ dictionary atoms.
The dictionary matrix $\bm{D}\in\mathbb{R}^{m\times d}$
and sparse coefficient matrix $\bm{C}\in\mathbb{R}^{d\times n}$ are found
by solving
\begin{equation}\label{Eq.CDL_0}
\mathop{\textup{minimize}}_{\bm{D}\in\mathcal{S}_D,\bm{C}\in\mathcal{S}_C}\quad
\dfrac{1}{2}\Vert\bm{X}-\bm{D}\bm{C}\Vert_F^2,
\end{equation}
\begin{equation}\label{Eq.CDL_1}
\text{or} \qquad \mathop{\textup{minimize}}_{\bm{D}\in\mathcal{S}_D,\bm{C}}\quad
\dfrac{1}{2}\Vert\bm{X}-\bm{D}\bm{C}\Vert_F^2+\lambda\Vert \bm{C}\Vert_1,
\end{equation}
where $\mathcal{S}_D:=\lbrace \bm{S}\in\mathbb{R}^{m\times d}:  \Vert\bm{s}_j\Vert \leq 1, \forall j=1,\ldots,d\rbrace$\footnote{Throughout this paper, given a matrix $\bm{X}$, we denote its $j$-th column by $\bm{x}_j$, and denote its entry at location $(i,j)$ by $x_{ij}$.} and $\mathcal{S}_C:=\lbrace \bm{S}\in\mathbb{R}^{d\times n}:  \Vert\bm{s}_j\Vert_0\leq k, \forall j=1,\ldots,n\rbrace$.

Formulations \eqref{Eq.CDL_0} and \eqref{Eq.CDL_1} cannot handle sparse noise and outliers.
Consequently, several robust dictionary learning (RDL) algorithms \cite{lu2013online,chen2013robust,wang2013semi,jiang2015robust,rdlhskf,8031891} have been proposed.
For instance, \cite{lu2013online} proposed to solve
\begin{equation}\label{Eq.RCDL_1}
\mathop{\textup{minimize}}_{\bm{D}\in\mathcal{S}_D,\bm{C}} \quad
\dfrac{1}{2}\Vert\bm{X}-\bm{D}\bm{C}\Vert_1+\lambda\Vert \bm{C}\Vert_1,
\end{equation}
using both batch and online optimization approaches.
In \cite{jiang2015robust}, the authors proposed to set threshold $\varepsilon$ and solve
\begin{equation}\label{Eq.RCDL_2}
\mathop{\textup{minimize}}_{\bm{D}\in\mathcal{S}_D,\bm{C}}\quad
\sum_{j=1}^n\min(\Vert\bm{x}_j-\bm{D}\bm{c}_j\Vert_2,\varepsilon)+\lambda\Vert \bm{C}\Vert_1,
\end{equation}
to limit the contribution of outliers to the objective.

Problem \eqref{Eq.RCDL_1} may fail
when the data are corrupted by small dense noise.
Problem \eqref{Eq.RCDL_2} finds outliers but cannot recover the clean data
because the hard-thresholding parameter $\varepsilon$ eliminates the representation loss for the outliers and then sets the corresponding columns of $\bm{C}$ to zero.

In contrast,
the following RDL formulation extended from \eqref{Eq.RCDL_1} \cite{lu2013online} simultaneously learns the dictionary and identifies the noise:
\begin{equation}\label{Eq.RCDL_3}
\mathop{\textup{minimize}}_{\bm{D}\in\mathcal{S}_D,\bm{C},\bm{E}}~
\dfrac{1}{2}\Vert\hat{\bm{X}}-\bm{D}\bm{C}-\bm{E}\Vert_F^2+\lambda_C\Vert \bm{C}\Vert_1+\lambda_E\mathcal{R}(\bm{E}).
\end{equation}
Here $\hat{\bm{X}}$ is the observed noisy matrix, $\bm{E}$ models the noise or outliers,
and $\mathcal{R}(\bm{E})$ penalizes  errors $\bm{E}$.
For example, when $\hat{\bm{X}}$ contains sparse noise, we set $\mathcal{R}(\bm{E})=\Vert\bm{E}\Vert_1$. The $\bm{D}$ obtained from \eqref{Eq.RCDL_3} can be used to denoise new data efficiently.

The formulation \eqref{Eq.RCDL_3} is closely related to a few other proposals.
For example, by setting $\bm{D}=\hat{\bm{X}}$,
one derives the self-expressive model used in sparse subspace clustering (SSC)  \cite{SSC_PAMIN_2013}.
Moreover, replacing $\Vert\bm{C}\Vert_1$ with  $\Vert\bm{C}\Vert_\ast$, one gets the low-rank representation (LRR) \cite{LRR_PAMI_2013} model.
A few extensions of SSC and LRR for subspace clustering can be found in \cite{KSSC,NIPS2013_4865, pmlr-v51-shen16,8573150}.
We may use SSC and LRR to remove additive noise and outliers.  For instance, when a few columns of $\bm{X}$ are outliers, SSC and LRR  can identify the outliers by encouraging $\bm{E}$  to be column-wise sparse. When $\bm{X}$ is corrupted by sparse noise,  we set $\mathcal{R}(\bm{E})=\Vert\bm{E}\Vert_1$. The recovered matrix is $\hat{\bm{X}}-\bm{E}$. Compared to RDL \eqref{Eq.RCDL_3}, in terms of denoising, the major advantage of SSC and LRR is that they are nonconvex. However, since the dictionary used in SSC and LRR is the noisy data matrix, the denoising performance may degrade. In addition, SSC and LRR are not effective in denoising new data.

\subsection{Kernel dictionary learning}

In recent years, several authors have proposed to
augment dictionary learning with kernel methods
to learn nonlinear structures
\cite{van2013design,7299018,liu2015robust,quan2016equiangular,fan2019online}.
For instance, \cite{van2013design} proposed to solve
\begin{equation}\label{Eq.KDL_13}
\begin{aligned}
\mathop{\textup{minimize}}_{\bm{D}\in\breve{\mathcal{S}}_D,\bm{C}\in\mathcal{S}_C}~
&\dfrac{1}{2}\Vert\phi(\hat{\bm{X}})-\phi(\hat{\bm{X}})\bm{D}\bm{C}-\phi(\hat{\bm{X}})\textup{diag}(\bm{w})\Vert_F^2\\
&+\lambda_C\Vert\bm{C}\Vert_0+\lambda_w\Vert\bm{w}\Vert_0,
\end{aligned}
\end{equation}
where $\phi$ denotes the feature map induced by a kernel function and $\breve{\mathcal{S}}_D:=\lbrace \bm{S}\in\mathbb{R}^{n\times d}:  \Vert\bm{s}_j\Vert=1, \forall j=1,2,\ldots,d\rbrace$.
The nonzeros of $\bm{w}$ in \eqref{Eq.KDL_13} identify the outliers in $\hat{\bm{X}}$.
In \cite{quan2016equiangular}, the following problem was considered:
\begin{equation}\label{Eq.KDL_16}
\mathop{\textup{minimize}}_{\bm{D}\in\bar{\mathcal{S}}_D,\bm{C}\in\mathcal{S}_C}\quad
\dfrac{1}{2}\Vert\phi(\hat{\bm{X}})-\phi(\bm{D})\bm{C}\Vert_F^2,
\end{equation}
where $\bar{\mathcal{S}}_D:=\lbrace \bm{S}\in\mathbb{R}^{m\times d}:  \Vert\bm{s}_j\Vert=1, \forall j=1,2,\ldots,d;\ \bm{s}_i^\top\bm{s}_j=\mu, \forall i\neq j\rbrace$, and $\mu$ is a predefined parameter.

One advantage of \eqref{Eq.KDL_16} over \eqref{Eq.KDL_13} is that the computational cost was reduced if $n\gg m$. Nevertheless, \eqref{Eq.KDL_16} is vulnerable to outliers. Moreover, neither \eqref{Eq.KDL_13} nor \eqref{Eq.KDL_16} can identify sparse noise in $\hat{\bm{X}}$ and recover the clean data. The reason is that the denoised matrix itself does not appear in the objective functions.

\subsection{Contributions of this paper} Our contributions are three-fold. First, we propose a new robust nonlinear matrix factorization model together with an effective optimization algorithm that explicitly separates the sparse noise or outliers from the observed data.
Second, we provide theory to prove correctness of our factorization in the feature space induced by kernels,
and justify the use of squared Frobenius norm regularization
on the feature matrix and coefficient matrix in the factorization model.
Finally, based on the robust nonlinear matrix factorization model, we propose a new subspace clustering method.
Extensive experiments on synthetic data, real image data, and real motion capture data showed that our proposed methods are more effective than baseline methods in dictionary learning, denoising, and subspace clustering.

\section{Robust Non-Linear Matrix Factorization}
\subsection{Low-rank factorization in kernel feature space}
Suppose the columns of a data matrix $\bm{X}\in\mathbb{R}^{m\times n}$ come from a generative model $\mathcal{M}$.
Let $\phi:\mathbb{R}^m\rightarrow\mathbb{R}^l$ be the feature map induced by a kernel function
\begin{equation*}
\mathcal{K}(\bm{x}_i,\bm{x}_j)=\langle \phi(\bm{x}_i),\phi(\bm{x}_j)\rangle=\phi(\bm{x}_i)^\top\phi(\bm{x}_j).
\end{equation*}
Two popular kernels are the polynomial (Poly) kernel and Gaussian radial basis function (RBF) kernel
\begin{equation*}
\begin{aligned}
&\mbox{Poly}:\ \mathcal{K}_{c,q}(\bm{x}_i,\bm{x}_j)=(\bm{x}_i^\top\bm{x}_j+c)^q\\
&\mbox{RBF}:\ \mathcal{K}_\sigma(\bm{x}_i,\bm{x}_j)=\exp\left(-\tfrac{1}{\sigma^2}\Vert \bm{x}_i-\bm{x}_j\Vert^2\right),
\end{aligned}
\end{equation*}
where $c$, $q$, and $\sigma$ are hyper-parameters.

Let $\phi(\bm{X})=[\phi(\bm{x}_1),\phi(\bm{x}_2),\ldots,\phi(\bm{x}_n)]$.
When $\phi(\bm{X})$ is low-rank or can be well approximated by a low-rank matrix,
we will seek a factorization of the form
\begin{equation}\label{Eq.phiFact}
\phi(\bm{X})\simeq\phi(\bm{D})\bm{C},
\end{equation}
where $\bm{D}\in\mathbb{R}^{m\times d}$ is a dictionary of $d$ atoms, $\bm{C}\in\mathbb{R}^{d\times n}$ is the coefficient matrix, and $d<\min\lbrace l,n\rbrace$.
This factorization model was also considered in \cite{van2013design,quan2016equiangular,fan2019online}.

We denote the feature map of the polynomial kernel by $\phi_{c,q}$. Then $\phi_{c,q}(\bm{X})\in\mathbb{R}^{\binom{m+q}{q}\times n}$. In \cite{fan2019online}, the authors assumed that the data generating model $\mathcal{M}$ is a union of $p$-degree polynomials with random coefficients, $f^{\lbrace j\rbrace}:\mathbb{R}^{r}\rightarrow\mathbb{R}^m$, $j=1,\ldots,k$, $r\ll m$; for $j=1,\ldots,k$, the $n/k$ columns of $\bm{X}$ are given by $\bm{x}=f^{\lbrace j\rbrace}(\bm{z})$, and $\bm{z}\in\mathbb{R}^{r}$ consists of $r$ uncorrelated random variables; see the following formulation
\begin{equation*}
\begin{aligned}
\bm{X}=[&f^{\lbrace 1\rbrace}(\bm{z}_1),\ldots,f^{\lbrace 1\rbrace}(\bm{z}_{n/k}),\\
&f^{\lbrace 2\rbrace}(\bm{z}_{n/k+1}),\ldots,f^{\lbrace 2\rbrace}(\bm{z}_{2n/k}),\ldots,\\
&f^{\lbrace k\rbrace}(\bm{z}_{n-n/k+1}), \ldots,f^{\lbrace k\rbrace}(\bm{z}_{n})]\bm{P},
\end{aligned}
\end{equation*}
where $\bm{P}\in\mathbb{R}^{n\times n}$ is an unknown permutation matrix.
They showed that
\begin{equation}
\textup{rank}(\bm{X})=\min\lbrace m,n,k\tbinom{r+p}{p}\rbrace,
\end{equation}
and
\begin{equation}\label{Eq.rankpolyk}
\textup{rank}(\phi_{c,q}(\bm{X}))=\min\lbrace \tbinom{m+q}{q},n,k\tbinom{r+pq}{pq}\rbrace.
\end{equation}
Thus, when  $p$ is large, $\bm{X}$ is high rank; but $\phi_{c,q}(\bm{X})$ is low-rank provided that $n$ is large enough.

Let $\phi_\sigma$ be the feature map of the Gaussian RBF kernel.
The following  reveals the connection between two types of kernels.
\begin{lemma}\label{lem_kk}
Define $s_{ij}:=\exp\left(-\tfrac{\Vert \bm{x}_i\Vert^2+\Vert\bm{x}_j\Vert^2+2c}{2\sigma^2}\right)$. Then for any $c\geq 0$, $\sigma$, $\bm{x}_i$, and $\bm{x}_j$,
\begin{equation*}
\begin{aligned}
\mathcal{K}_\sigma(\bm{x}_i,\bm{x}_j)=s_{ij}\sum_{u=0}^\infty\dfrac{\mathcal{K}_{c,u}(\bm{x}_i,\bm{x}_j)}{\sigma^{2u}u!}.
\end{aligned}
\end{equation*}
\end{lemma}
It follows from Lemma \ref{lem_kk} that
$$
\phi_\sigma(\bm{x}_i)=s_i \left[w_0\phi_{c,0}^\top(\bm{x}_i),
w_1\phi_{c,1}^\top(\bm{x}_i), \ldots, w_\infty\phi_{c,\infty}^\top(\bm{x}_i)\right]^\top,
$$
where $s_i=\exp\left(-\tfrac{\Vert \bm{x}_i\Vert^2+c}{2\sigma^2}\right)$ and $w_u=1/(\sigma^u\sqrt{u!})$ for $u=0,1,\ldots,\infty$. Therefore, $\phi_\sigma(\bm{X})\in\mathbb{R}^{\infty\times n}$ is full rank, according to \eqref{Eq.rankpolyk}. Let $\bm{S}_X=\textup{diag}(s_1,\ldots,s_n)$ and $\bm{S}_D=\textup{diag}\left(\exp(-\tfrac{\Vert \bm{d}_1\Vert^2+c}{2\sigma^2}),\ldots,\exp(-\tfrac{\Vert \bm{d}_d\Vert^2+c}{2\sigma^2})\right)$. We have
\begin{lemma}\label{lem_objk1k2}
Define $\kappa_1=\max\lbrace0.5n,\sqrt{dn}\Vert\bm{C}\Vert_F,$ $0.5d\Vert\bm{C}\Vert_2\Vert\bm{C}\Vert_F\rbrace$ and $\kappa_2=\scaleobj{0.9}{\max\lbrace\max_{i}\Vert\bm{x}_i\Vert^2,\max_{j}\Vert\bm{d}_j\Vert^2\rbrace}$. Suppose $\sigma^2>\kappa_2+c$. Then for any $q\geq 0$,
\begin{equation*}
\begin{aligned}
&\dfrac{1}{2}\Vert\phi_\sigma(\bm{X})-\phi_\sigma(\bm{D})\bm{C}\Vert_F^2\\
=&\sum_{u=0}^{q}\dfrac{w_u^2}{2}\Vert\phi_{c,u}(\bm{X})\bm{S}_X-\phi_{c,u}(\bm{D})\bm{S}_D\bm{C}\Vert_F^2+R,
\end{aligned}
\end{equation*}
where $\vert R\vert\leq \dfrac{3\kappa_1\exp(-\tfrac{c}{\sigma^2})}{q!}\left(\dfrac{\kappa_2+c}{\sigma^2}\right)^{q}$.
\end{lemma}

Lemma \ref{lem_objk1k2} shows the factorization error of the feature matrix induced by the RBF kernel
can be well approximated by the weighted sum of the factorization errors
of the feature matrices induced by polynomial kernels of different degrees
if $\sigma$ and $q$ are sufficiently large.
Notice that
$\textup{rank}(\phi_{c,u}(\bm{X})\bm{S}_X)=\textup{rank}(\phi_{c,u}(\bm{X}))$
because $\bm{S}_X$ is diagonal with positive diagonal entries.
The following corollary of Lemma \ref{lem_objk1k2} provides an upper bound on
the factorization error using the RBF kernel:
\begin{corollary}\label{cor_Ebound}
Suppose $d\geq\textup{rank}(\phi_{c,q}(\bm{X}))$ and $\sigma^2>\kappa_2+c$. Then for any $q\geq 1$, there exist $\bm{D}$ and $\bm{C}$ such that
\begin{equation*}
\dfrac{1}{2}\Vert\phi_\sigma(\bm{X})-\phi_\sigma(\bm{D})\bm{C}\Vert_F^2\leq\dfrac{3\kappa_1\exp(-\tfrac{c}{\sigma^2})}{q!}\left(\dfrac{\kappa_2+c}{\sigma^2}\right)^{q}.
\end{equation*}
\end{corollary}
\begin{remark}
\textup{Since $\kappa_1$ relates to $\Vert\bm{C}\Vert_F$, the bound in Corollary \ref{cor_Ebound} may be tighter when $\Vert\bm{C}\Vert_F^2$ is smaller.}
\end{remark}
Hence, when $d\geq k\tbinom{r+pq}{pq}$ and $n$, $\sigma$ are sufficiently large, our factorization model \eqref{Eq.phiFact} holds as follows:
\begin{align*}
\phi_{c,q}(\bm{X})=\phi_{c,q}(\bm{D})\bm{C}\quad \textup{and}\quad \phi_{\sigma}(\bm{X})\approx\phi_{\sigma}(\bm{D})\bm{C}.
\end{align*}
When using polynomial kernel or Gaussian RBF kernel, we can exactly or approximately factorize $\phi(\bm{X})$ into $\phi(\bm{D})$ and $\bm{C}$, where $d$ could be much smaller than $n$. This property enables us to extract nonlinear features (rows of $\bm{C}$), find useful basis elements (columns of $\bm{D}$), or remove noise from $\bm{X}$.

\subsection{Robustness in data space}
\subsubsection{General objective function}
Suppose a data matrix $\bm{X}\in\mathbb{R}^{m\times n}$ is partially corrupted
by sparse noise $\bm{E}\in\mathbb{R}^{m\times n}$ to form noisy observations
\begin{equation}\label{Eq.XE}
\hat{\bm{X}}=\bm{X}+\bm{E},
\end{equation}
where the locations of nonzero entries of $\bm{E}$ are uniform and random.
We wish to recover $\bm{X}$ from $\hat{\bm{X}}$.
Using \eqref{Eq.phiFact} and \eqref{Eq.XE}, we define the factorization loss as
\begin{align}
&\mathcal{L}(\bm{D},\bm{C},\bm{E}):=\dfrac{1}{2}\Vert\phi(\hat{\bm{X}}-\bm{E})-\phi(\bm{D})\bm{C}\Vert_F^2 \label{Eq.Ldef}\\
=&\scaleobj{0.95}{\dfrac{1}{2}\textup{Tr}\left(\phi(\hat{\bm{X}}-\bm{E})^{\top}\phi(\hat{\bm{X}}-\bm{E})\right)-\textup{Tr}\left(\bm{C}^{\top}\phi(\bm{D})^{\top}\phi(\hat{\bm{X}}-\bm{E})\right)} \nonumber\\
&\scaleobj{0.95}{+\dfrac{1}{2}\textup{Tr}\left(\bm{C}^{\top}\phi(\bm{D})^{\top}\phi(\bm{D})\bm{C}\right)}, \nonumber
\end{align}
where $\textup{Tr}(\cdot)$ denotes the matrix trace.
Using the kernel,
we have $\phi(\hat{\bm{X}}-\bm{E})^{\top}\phi(\hat{\bm{X}}-\bm{E})=\mathcal{K}(\hat{\bm{X}}-\bm{E},\hat{\bm{X}}-\bm{E})\in\mathbb{R}^{n\times n}$,
$\phi(\bm{D})^{\top}\phi(\hat{\bm{X}}-\bm{E})=\mathcal{K}(\bm{D},\hat{\bm{X}}-\bm{E})\in\mathbb{R}^{d\times n}$,
and $\phi(\bm{D})^{\top}\phi(\bm{D})=\mathcal{K}(\bm{D},\bm{D})\in\mathbb{R}^{d\times d}$.
Hence from \eqref{Eq.Ldef},
\begin{align*}
&\mathcal{L}(\bm{D},\bm{C},\bm{E})
=\dfrac{1}{2}\textup{Tr}\left(\mathcal{K}(\hat{\bm{X}}-\bm{E},\hat{\bm{X}}-\bm{E})\right)\\
&-\textup{Tr}\left(\bm{C}^{\top}\mathcal{K}(\bm{D},\hat{\bm{X}}-\bm{E})\right)
+\dfrac{1}{2}\textup{Tr}\Big(\bm{C}^{\top}\mathcal{K}(\bm{D},\bm{D})\bm{C}\Big).
\end{align*}

In addition, we define the regularization as
\begin{equation*}
\begin{aligned}
&\mathcal{R}(\bm{D},\bm{C},\bm{E}):=\lambda_{D}\mathcal{R}(\bm{D})+\lambda_{C}\mathcal{R}(\bm{C})+\lambda_{E}\mathcal{R}(\bm{E}),
\end{aligned}
\end{equation*}
where $\lambda_{D}$, $\lambda_{C}$, and $\lambda_{E}$ are penalty parameters.
Then we propose to solve
\begin{equation}\label{Eq.RNLMF_1}
\mathop{\textup{minimize}}_{\bm{D},\bm{C},\bm{E}}\ \ \mathcal{L}(\bm{D},\bm{C},\bm{E})+\mathcal{R}(\bm{D},\bm{C},\bm{E})\triangleq\mathcal{J}(\bm{D},\bm{C},\bm{E}).
\end{equation}
\begin{remark}
\textup{In \eqref{Eq.RNLMF_1}, we can introduce a constraint $\bm{D}=\hat{\bm{X}}-\bm{E}$ to eliminate $\bm{D}$, which leads to a self-expressive model, i.e. represent $\phi(\hat{\bm{X}}-\bm{E})$ with itself multiplied by $\bm{C}$; then we only need to compute $\bm{C}$ and $\bm{E}$. However, as $\bm{C}\in\mathbb{R}^{n\times n}$, the space and time complexities will increase significantly.}
\end{remark}

\subsubsection{Noise-specific penalty on $\bm{E}$}
We suggest choosing penalties of $\bm{E}$ from below, based on the suspected noise distribution.
\begin{itemize}
\item When all entries of $\bm{X}$ are corrupted by Gaussian noise, i.e. $E_{ij}\sim\mathcal{N}(0,\epsilon^2)$, $\forall$ $(i,j)\in[m]\times[n]$,  we set $\mathcal{R}(\bm{E})=\tfrac{1}{2}\Vert\bm{E}\Vert_F^2$.
\item When $\bm{X}$ is partially and randomly corrupted, i.e., $\bm{E}$ is a sparse matrix and $\mathbb{P}[E_{ij}\neq 0]=\rho$, we set  $\mathcal{R}(\bm{E})=\Vert\bm{E}\Vert_1$, where $0<\rho<1$ and $\Vert\cdot\Vert_1$ denotes the $\ell_1$ norm of vector or matrix serving as a convex relaxation of the $\ell_0$ norm.
\item When a few columns of $\bm{X}$ are corrupted by Gaussian noise, we set $\mathcal{R}(\bm{E})=\Vert\bm{E}\Vert_{2,1}$, where $\Vert\bm{E}\Vert_{2,1}=\sum_{j=1}^n\Vert\bm{e}_{j}\Vert_2$ is a convex relaxation of the number of nonzero columns of $\bm{E}$.
\end{itemize}
Next, we detail the choices of $\mathcal{R}(\bm{D})$ and $\mathcal{R}(\bm{C})$.

\subsubsection{Smooth penalty on $\bm{D}$ and $\bm{C}$}
We can set $\mathcal{R}(\bm{D})=\tfrac{1}{2}\Vert\phi(\bm{D})\Vert_F^2$ and $\mathcal{R}(\bm{C})=\tfrac{1}{2}\Vert\bm{C}\Vert_F^2$. We have $\mathcal{R}(\bm{D})=\tfrac{1}{2}\textup{Tr}\left( \phi(\bm{D})^{\top}\phi(\bm{D}) \right)=\tfrac{1}{2}\textup{Tr}\left(\mathcal{K}(\bm{D},\bm{D})\right)$. In this case, we often require that $d$ is equal to (or a little bit larger than) the rank or approximate rank of $\phi(\bm{X})$. Otherwise, the recovery error will be large.

\begin{lemma}\label{lem_lbound_0}
Let $\bm{X}\in\mathbb{R}^{m\times n}$, $\bm{D}\in\mathbb{R}^{m\times d}$, and $\bm{C}\in\mathbb{R}^{d\times n}$.  For any $\phi:\mathbb{R}^m\rightarrow\mathbb{R}^l$, suppose $d\geq \textup{rank}(\phi(\bm{X}))$, then
\begin{equation*}
\min_{\phi(\bm{D})\bm{C}=\phi(\bm{X})}\dfrac{1}{2}\Vert\phi(\bm{D})\Vert_F^2+\dfrac{1}{2}\Vert\bm{C}\Vert_F^2\geq\Vert \phi(\bm{X})\Vert_\ast.
\end{equation*}
\end{lemma}
Lemma \ref{lem_lbound_0} shows using factorization we can minimize an upper bound of $\Vert \phi(\bm{X})\Vert_\ast$, which may eventually reduce the value of $\Vert \phi(\bm{X})\Vert_\ast$. The following corollary implies that solving problem \eqref{Eq.RNLMF_1} may find a low-nuclear-norm matrix $\phi(\bm{D})\bm{C}$ to approximate $\phi(\hat{\bm{X}}-\bm{E})$.
\begin{corollary}\label{cor_boundDC}
For any $\phi:\mathbb{R}^m\rightarrow\mathbb{R}^l$,
\begin{equation*}
\dfrac{\lambda_D}{2}\Vert\phi(\bm{D})\Vert_F^2+\dfrac{\lambda_C}{2}\Vert\bm{C}\Vert_F^2\geq \sqrt{\lambda_C\lambda_D}\Vert \phi(\bm{D})\bm{C}\Vert_\ast.
\end{equation*}
\end{corollary}
Nevertheless, we may not achieve the equality in Lemma \ref{lem_lbound_0} and Corollary \ref{cor_boundDC} because of the presence of $\phi$.
Notice that with RBF kernel, $\textup{Tr}\big(\mathcal{K}(\bm{D},\bm{D})\big)\equiv d$, which means $\lambda_{D}\mathcal{R}(\bm{D})$ has no effect on the minimization and can be discarded; thus the number of penalty parameters is reduced. The following lemma explains the connection between $\Vert\bm{C}\Vert_F^2$ and $\Vert\phi(\bm{D})\bm{C}\Vert_\ast$ when using RBF kernel.
\begin{lemma}\label{cor_rbfFbound}
Suppose $\phi$ is induced by RBF kernel. Then for any $\bm{D}\in\mathbb{R}^{m\times d}$ and $\bm{C}\in\mathbb{R}^{d\times n}$,
\begin{equation*}
\Vert\bm{C}\Vert_F\geq \Vert\phi(\bm{D})\bm{C}\Vert_\ast/\sqrt{d}.
\end{equation*}
\end{lemma}

\subsubsection{Non-smooth penalty on $\bm{D}$ and $\bm{C}$}
For example, we set $\mathcal{R}(\bm{D})=\Vert\phi(\bm{D})\Vert_\ast$, where $\Vert\cdot\Vert_\ast$ denotes the matrix nuclear norm. We have $\mathcal{R}(\bm{D})=\textup{Tr}\big(\mathcal{K}(\bm{D},\bm{D})^{1/2}\big)$.
Such penalty on $\bm{D}$ will encourage $\phi(\bm{D})$ to be low-rank.
Similarly, we can penalize $\bm{C}$ to be low-rank by $\mathcal{R}(\bm{C})=\Vert\bm{C}\Vert_\ast$.
Moreover, we may set $\mathcal{R}(\bm{C})=\Vert\bm{C}\Vert_1$, which encourages $\bm{C}$ to be sparse.
The motivation is the same as dictionary learning: each column of $\phi(\bm{X})$ can be represented by a linear combination of a few columns of $\phi(\bm{D})$. Nevertheless, the nonsmooth $\mathcal{R}(\bm{D})$ and $\mathcal{R}(\bm{C})$ will increase the difficulty of optimization.

In the remaining of this paper, we will focus on Gaussian RBF kernel because of the following reasons.
First, we only need to determine one parameter $\sigma$ in Gaussian RBF kernel, compared to two parameters $c$ and $q$ in polynomial kernel.
Second, Gaussian RBF kernel is easier to optimize and the parameter $\sigma$ controls the weights of low-order features and high-order features effectively.
In addition, as mentioned above, when using Gaussian RBF kernel and $\mathcal{R}(\bm{D})=\tfrac{1}{2}\Vert\phi(\bm{X})\Vert_F^2$, we do not need the parameter $\lambda_D$.

\section{Optimization for RNLMF}
Problem \eqref{Eq.RNLMF_1} is nonconvex and nonsmooth and has three blocks of variables. We propose to initialize $\bm{D}$ randomly and initialize $\bm{E}$ with zeros, then update $\bm{C}$, $\bm{D}$, and $\bm{E}$ alternately. In the numerical results, we show that the alternating minimization always provide satisfactory denoising performance, provided that parameters such as $\lambda_E$ are properly determined.

\subsection{Update $\bm{C}$ by closed-form solution or proximal gradient method}\label{Sec.update_D}
Fix $\bm{D}$ and $\bm{E}$ and let
\begin{equation*}
\begin{aligned}
\mathcal{L}(\bm{C})=&-\textup{Tr}\left(\bm{C}^{\top}\mathcal{K}(\bm{D}_{t-1},\hat{\bm{X}}-\bm{E}_{t-1})\right)\\
&+\dfrac{1}{2}\textup{Tr}\Big(\bm{C}^{\top}\mathcal{K}(\bm{D}_{t-1},\bm{D}_{t-1})\bm{C}\Big).
\end{aligned}
\end{equation*}
We aim to solve
\begin{equation}\label{Eq.prob_C}
\mathop{\text{minimize}}_{\bm{C}}\quad\mathcal{L}(\bm{C})+\lambda_{C}\mathcal{R}(\bm{C}).
\end{equation}

When $\mathcal{R}(\bm{C})=\tfrac{1}{2}\Vert C\Vert_F^2$, by letting $\nabla_{\bm{C}}[\mathcal{L}(\bm{C})+\lambda_{C}\mathcal{R}(\bm{C})]=0$, we update $\bm{C}$ to the solution of \eqref{Eq.prob_C}:
\begin{equation}\label{Eq.C_sol_1}
\bm{C}_{t}
=(\mathcal{K}(\bm{D}_{t-1},\bm{D}_{t-1})+\lambda_{C} \bm{I}_d)^{-1}\mathcal{K}(\bm{D}_{t-1},\hat{\bm{X}}-\bm{E}_{t-1}),
\end{equation}
where $\bm{I}_d\in\mathbb{R}^{d\times d}$ is an identity matrix. The problem is actually closely related to the kernel ridge regression, in which the feature map is performed only on the regressors $\bm{D}$ and $\phi(\hat{\bm{X}}-\bm{E})$ is replaced by the dependent variables.

When $\mathcal{R}(\bm{C})=\Vert C\Vert_1$ or $\Vert C\Vert_\ast$, \eqref{Eq.prob_C} has no closed-form solution. We use first order approximation to find a majorizer of $\mathcal{L}(\bm{C})$ at $\bm{C}_{t-1}$
$$\mathcal{L}(\bm{C})\leq \mathcal{L}(\bm{C}_{t-1})+\langle \nabla_{\bm{C}}\mathcal{L}(\bm{C}_{t-1}), \bm{C}-\bm{C}_{t-1}\rangle+\dfrac{\tau_C^t}{2}\Vert\bm{C}-\bm{C}_{t-1}\Vert_F^2$$
and then solve
\begin{equation}\label{Eq.prob_C_L1}
\mathop{\text{minimize}}_{\bm{C}}~\dfrac{\tau_C^t}{2}\Vert\bm{C}-\bm{C}_{t-1}+\nabla_{\bm{C}}\mathcal{L}(\bm{C}_{t-1})/\tau^t_C\Vert_F^2+\lambda_{C}\mathcal{R}(\bm{C}),
\end{equation}
where $\nabla_{\bm{C}}\mathcal{L}(\bm{C}_{t-1})=-\mathcal{K}(\bm{D}_{t-1},\hat{\bm{X}}-\bm{E}_{t-1})+\mathcal{K}(\bm{D}_{t-1},\bm{D}_{t-1})\bm{C}_{t-1}$.
Here we need $\tau_C^t>\Vert \mathcal{K}(\bm{D}_{t-1},\bm{D}_{t-1})\Vert_2$ to ensure that \eqref{Eq.prob_C_L1} is non-expansive. Consequently, the closed-form solution of \eqref{Eq.prob_C_L1} as well as \eqref{Eq.C_sol_1} are shown in Table \ref{Tab_sol_C}.
In the table, $\Theta$ denotes the soft-thresholding operator defined by
\begin{equation*}
\Theta_{u}(v)=\dfrac{\vert v\vert}{v}\max(\vert v\vert-u,0).
\end{equation*}
\noindent In addition, $\Psi$ denotes the singular value thresholding operator \cite{CaiCandesShen2010} defined by
\begin{equation*}
\Psi_{u}(\bm{M})=\bm{U}\Theta_u(\bm{S})\bm{V}^{\top},
\end{equation*}
where $\bm{U}$, $\bm{S}$, and $\bm{V}$ are given by the SVD $\bm{M}=\bm{U}\bm{S}\bm{V}^{\top}$.

\begin{table}[h!]
\centering
\caption{Solution for $\bm{C}_t$ with respect to different $\mathcal{R}(\bm{C})$}
\begin{tabular}{|c|c|}
\hline
$\mathcal{R}(\bm{C})$ & $\bm{C}_{t}$ \\  \hline
$\tfrac{1}{2}\Vert \bm{C}\Vert_F^2$ &  {\footnotesize $(\mathcal{K}(\bm{D}_{t-1},\bm{D}_{t-1})+\lambda_{C} \bm{I}_d)^{-1}\mathcal{K}(\bm{D}_{t-1},\hat{\bm{X}}-\bm{E}_{t-1})$}\\
$\Vert \bm{C}\Vert_1$ & $\Theta_{\lambda_C/\tau_C^t}\big(\bm{C}_{t-1}-\nabla_{\bm{C}}\mathcal{L}(\bm{C}_{t-1})/\tau_C^t\big)$ \\
$\Vert \bm{C}\Vert_\ast$ & $\Psi_{\lambda_C/\tau_C^t}\big(\bm{C}_{t-1}-\nabla_{\bm{C}}\mathcal{L}(\bm{C}_{t-1})/\tau_C^t\big)$ \\ \hline
\end{tabular}\label{Tab_sol_C}
\end{table}

The following lemma shows that the update of $\bm{C}$ when $\mathcal{R}(\bm{C})=\Vert \bm{C}\Vert_1$ or $\Vert \bm{C}\Vert_\ast$ ensures the objective function is nonascending.
\begin{lemma}\label{lem_LC}
Let $\bm{C}_t$ be the solution of \eqref{Eq.prob_C_L1} with $\mathcal{R}(\bm{C})=\Vert \bm{C}\Vert_1$ or $\Vert \bm{C}\Vert_\ast$. Denote  $L_C^t=\Vert \mathcal{K}(\bm{D}_{t-1},\bm{D}_{t-1})\Vert_2$. Then
\begin{equation*}
\begin{aligned}
&\mathcal{J}(\bm{D}_{t-1},\bm{C}_{t},\bm{E}_{t-1})-\mathcal{J}(\bm{D}_{t-1},\bm{C}_{t-1},\bm{E}_{t-1})\\
\leq &-\dfrac{\tau_C^t-L_C^t}{2}\Vert \bm{C}_t-\bm{C}_{t-1}\Vert_F^2,
\end{aligned}
\end{equation*}
where $\tau_C^t>L_C^t$.
\end{lemma}

\subsection{Update $\bm{D}$ by relaxed Newton method}
Fix $\bm{C}$ and $\bm{E}$ and let
\begin{equation*}
\begin{aligned}
\mathcal{L}(\bm{D})=&-\textup{Tr}\left(\bm{C}_t^{\top}\mathcal{K}(\bm{D},\hat{\bm{X}}-\bm{E}_{t-1})\right)\\
&+\dfrac{1}{2}\textup{Tr}\Big(\bm{C}_t^{\top}\mathcal{K}(\bm{D},\bm{D})\bm{C}_t\Big).
\end{aligned}
\end{equation*}
Because of the presence of kernel function, the minimization of $\mathcal{L}(\bm{D})$ has no closed-form solution.  The gradient\footnote{Use the chain rule
$\tfrac{\partial{\mathcal{L}}}{\partial{\bm{Z}}}=\sum_{i=1}^{n_1}\sum_{j=1}^{n_2}\tfrac{\partial{\mathcal{L}}}{\partial{\mathcal{K}_{ij}}}\tfrac{\partial{\mathcal{K}_{ij}}}{\partial{\bm{Z}}}$, where $\mathcal{K}\in\mathbb{R}^{n_1\times n_2}$ is the kernel matrix computed from $\bm{Z}$. }
 is
\begin{equation*}
\begin{aligned}
\nabla_{\bm{D}}\mathcal{L}(\bm{D})=&\tfrac{1}{\sigma^2}((\hat{\bm{X}}-\bm{E}_{t-1})\bm{W}_1^t-\bm{D}\bar{\bm{W}}_1^t)\\
&+\tfrac{2}{\sigma^2}(\bm{D}\bm{W}_2^t-\bm{D}\bar{\bm{W}}_2^t),
\end{aligned}
\end{equation*}
where $\bm{W}_1^t=-\bm{C}_{t}^\top\odot\mathcal{K}(\hat{\bm{X}}-\bm{E}_{t-1},\bm{D})$, $\bm{W}_2^t=(0.5\bm{C}_t\bm{C}_t^\top)\odot\mathcal{K}(\bm{D},\bm{D})$, $\bar{\bm{W}}_1^t=\mbox{diag}(\bm{1}_n^\top\bm{W}_1^t)$, and $\bar{\bm{W}}_2=\mbox{diag}(\bm{1}_d^\top\bm{W}_2)$.
One straightforward approach is to update $\bm{D}$ by gradient descent with backtracking line search, which however requires evaluating $\mathcal{L}(\bm{D})$ for multiple times and hence  increases the computational cost. In addition, one may consider using second-order information to accelerate the optimization.
Note that $\tfrac{\partial[\bm{W}_1^t]_{ij}}{\partial [\bm{D}]_{:j}}=[\bm{C}_t^\top]_{ij}[\mathcal{K}(\hat{\bm{X}}-\bm{E}_{t-1},\bm{D})]_{ij}(\hat{\bm{x}_i}-[\bm{E}_{t-1}]_{:j})/\sigma^2$. Thus, when $\sigma$ is large, we regard $\bm{W}_1^t$ (also $\bar{\bm{W}}_1^t$, $\bm{W}_2^t$, and $\bar{\bm{W}}_2^t$) as a constant. Then we update $\bm{D}$ by a relaxed Newton step
\begin{equation}\label{Eq.D_sol_1}
\bm{D}_{t}=\bm{D}_{t-1}-\tfrac{1}{\tau_D^t}\nabla_{\bm{D}}\mathcal{L}(\bm{D}_{t-1})(\bm{H}_{t-1}+\mu\bm{I})^{-1},
\end{equation}
where $\bm{H}_{t-1}=\tfrac{1}{\sigma^2}(-\bar{\bm{W}}_1^t+2\bm{W}_2^t-2\bar{\bm{W}}_2^t)$, $\tau_D^t\geq 1$ controls the step size, and $\mu\geq 0$ is large enough such that $\bm{H}_{t-1}+\mu\bm{I}$ is positive definite.
The effectiveness of  \eqref{Eq.D_sol_1} is  justified by the following lemma.
\begin{lemma}\label{lem_LD}
Suppose $\bm{D}_t$ is given by \eqref{Eq.D_sol_1} and $\tau_D^t$ and $\mu$ are sufficiently large. Then
\begin{equation*}
\begin{aligned}
&\mathcal{J}(\bm{D}_{t},\bm{C}_{t},\bm{E}_{t-1})-\mathcal{J}(\bm{D}_{t-1},\bm{C}_{t},\bm{E}_{t-1})\\
\leq &-\tfrac{1}{2\tau_D^t}\textup{Tr}\left(\nabla_{\bm{D}}\mathcal{L}(\bm{D}_{t-1})(\bm{H}_{t-1}+\mu\bm{I})^{-1}\nabla_{\bm{D}}\mathcal{L}(\bm{D}_{t-1})^\top\right)\leq 0.\\
\end{aligned}
\end{equation*}
\end{lemma}
\begin{remark}
Empirically, in our experiments, we found $\bm{H}$ was always positive definite. Hence we set $\mu=0$ in all experiments. In addition, $\tau_D=1$ works well in practice.
\end{remark}

We can also incorporate momentum into the update of $\bm{D}$:
\begin{equation}\label{Eq.D_sol_delta}
\bm{\Delta}_t=\eta\bm{\Delta}_{t-1}+\tfrac{1}{\tau_D^t}\nabla_{\bm{D}}\mathcal{L}(\bm{D}_{t-1})(\bm{H}_{t-1}+\mu\bm{I})^{-1},
\end{equation}
 where $0<\eta<1$; then
\begin{equation}\label{Eq.D_sol_2}
\bm{D}_{t}=\bm{D}_{t-1}-\bm{\Delta}_t.
\end{equation}
The following corollary shows that when $\eta$ is sufficiently small, the objective function is non-increasing.

\begin{corollary}\label{cor_LD}
Suppose $\bm{D}_t$ is given by \eqref{Eq.D_sol_2} and $\tau_D^t$ is sufficiently large. Then
\begin{equation}
\begin{aligned}
&\mathcal{J}(\bm{D}_{t},\bm{C}_{t},\bm{E}_{t-1})-\mathcal{J}(\bm{D}_{t-1},\bm{C}_{t},\bm{E}_{t-1})\\
\leq &-\dfrac{1}{2\tau_D^t}\textup{Tr}\left(\nabla_{\bm{D}}\mathcal{L}(\bm{D}_{t-1})(\bm{H}_{t-1}+\mu\bm{I})^{-1}\nabla_{\bm{D}}\mathcal{L}(\bm{D}_{t-1})^\top\right)\\
&+\dfrac{\eta^2\tau_D^t}{2}\textup{Tr}\left(\bm{\Delta}_{t-1}(\bm{H}_{t-1}+\mu\bm{I})\bm{\Delta}_{t-1}^\top\right).
\end{aligned}
\end{equation}
\end{corollary}

When $d$ is large, to avoid the high computational cost of the inverse of $\bm{H}_{t-1}$, we suggest replacing \eqref{Eq.D_sol_2} with
\begin{equation}\label{Eq.D_sol_3}
\bm{D}_{t}=\bm{D}_{t-1}-\bar{\bm{\Delta}}_t,
\end{equation}
where $\bar{\bm{\Delta}}_t=\eta\bar{\bm{\Delta}}_{t-1}+\tfrac{1}{\tau_D^t}\nabla_{\bm{D}}\mathcal{L}(\bm{D}_{t-1})/\Vert\bm{H}_{t-1}\Vert_2$.

\renewcommand{\algorithmicrequire}{\textbf{Input:}}
\renewcommand{\algorithmicensure}{\textbf{Output:}}
\begin{algorithm}[h]
\caption{Optimization for RNLMF}
\label{alg.RNLMF}
\begin{algorithmic}[1]
\Require
$\hat{\bm{X}}$, $d$, $\lambda_C$, $\lambda_E$, $\sigma$, $\eta$, $t_{\text{iter}}$.
\State Initialize: $\bm{E}=\bm{0}$, $\bm{C}=\bm{0}$, $\bm{D}\sim\mathcal{N}(0,1)$, $\bm{\Delta}=\bm{0}$, $t=0$.
\Repeat
\State $t\leftarrow t+1$.
\State Update $\bm{C}$ using Table \ref{Tab_sol_C}.
\State Update $\bm{\Delta}$ using \eqref{Eq.D_sol_delta}
\State Update $\bm{D}$ using \eqref{Eq.D_sol_2}.
\State Update $\bm{E}$ using Table \ref{Tab_sol_E}.
\Until{converged or $t=t_{\text{iter}}$}
\Ensure $\bm{X}=\hat{\bm{X}}-\bm{E}$, $\bm{D}$, $\bm{C}$.
\end{algorithmic}
\end{algorithm}

\subsection{Update $\bm{E}$ by proximal gradient method}
Fix $\bm{C}$ and $\bm{D}$ and let
\begin{equation*}
\begin{aligned}
\mathcal{L}(\bm{E})=&\dfrac{1}{2}\textup{Tr}\left(\mathcal{K}(\hat{\bm{X}}-\bm{E},\hat{\bm{X}}-\bm{E})\right)-\textup{Tr}\left(\bm{C}_t^\top\mathcal{K}(\bm{D}_t,\hat{\bm{X}}-\bm{E})\right)\\
=&\dfrac{n}{2}-\textup{Tr}\left(\bm{C}_t^\top\mathcal{K}(\bm{D}_t,\hat{\bm{X}}-\bm{E})\right)\\
\end{aligned}
\end{equation*}
Then we need to solve
\begin{equation}\label{Eq.prob_E}
\mathop{\text{minimize}}_{\bm{E}}\quad\mathcal{L}(\bm{E})+\lambda_{E}\mathcal{R}(\bm{E}),
\end{equation}
which, however, has no closed-form solution. Compute the gradient of $\mathcal{L}(\bm{E})$ as
\begin{equation}\label{Eq.gradE}
\nabla_{\bm{E}}\mathcal{L}(\bm{E})=\tfrac{1}{\sigma^2}\big((\hat{\bm{X}}-\bm{E})\bar{\bm{W}}_3^t-\bm{D}_t\bm{W}_3^t\big),
\end{equation}
where $\bm{W}_3^t=-\bm{C}_{t}\odot\mathcal{K}(\bm{D}_t,\hat{\bm{X}}-\bm{E})$ and $\bar{\bm{W}}_3^t=\mbox{diag}(\bm{1}_d^\top\bm{W}_3^t)$.
Suppose the Lipschitz constant of $\nabla_{\bm{E}}\mathcal{L}(\bm{E})$ at iteration $t$ is $L_E^t$.
Let $\tau_E^t>L_E^t$  and we have
\begin{equation*}
\begin{aligned}
\mathcal{L}(\bm{E})\leq &\mathcal{L}(\bm{E}_{t-1})+\langle \nabla_{\bm{E}}\mathcal{L}(\bm{E}_{t-1}),\bm{E}-\bm{E}_{t-1}\rangle\\
&+\dfrac{\tau_E^t}{2}\Vert \bm{E}-\bm{E}_{t-1}\Vert_F^2.
\end{aligned}
\end{equation*}
Thus we update $\bm{E}$ by solving
\begin{equation}\label{Eq.prob_E1}
\mathop{\text{minimize}}_{\bm{E}}~\dfrac{\tau_E^t}{2}\Vert \bm{E}-\bm{E}_{t-1}+\nabla_{\bm{E}}\mathcal{L}(\bm{E}_{t-1})/\tau_E^t\Vert_F^2+\lambda_{E}\mathcal{R}(\bm{E}).
\end{equation}

The closed-form solutions of \eqref{Eq.prob_E1} with different $\mathcal{R}(\bm{E})$ are shown in Table \ref{Tab_sol_E}.
In the table, $\Upsilon_u(\cdot)$ is the column-wise soft-thresholding operator \cite{parikh2014proximal} defined as
\begin{equation*}
\Upsilon_u(\bm{v})=\left\{
\begin{array}{ll}
\tfrac{(\Vert\bm{v}\Vert-u)\bm{v}}{\Vert\bm{v}\Vert}, &\textup{if}\ \Vert\bm{v}\Vert>u;\\
\bm{0}, &\textup{otherwise.}
\end{array}
\right.
\end{equation*}

\begin{table}[h!]
\centering
\caption{Solution of \eqref{Eq.prob_E1} with respect to different $\mathcal{R}(\bm{E})$}
\begin{tabular}{|c|c|}
\hline
$\mathcal{R}(\bm{E})$ & $\bm{E}_{t}$ \\  \hline
$\tfrac{1}{2}\Vert \bm{E}\Vert_F^2$ & $\dfrac{\tau_E^t}{\tau_E^t+\lambda_E}(\bm{E}_{t-1}-\nabla_{\bm{E}}\mathcal{L}(\bm{E}_{t-1})/\tau_E^t)$ \\
$\Vert \bm{E}\Vert_1$ & $\Theta_{\lambda_E/\tau_E^t}(\bm{E}_{t-1}-\nabla_{\bm{E}}\mathcal{L}(\bm{E}_{t-1})/\tau_E^t)$ \\
$\Vert \bm{E}\Vert_{2,1}$ & $\Upsilon_{\lambda_E/\tau_E^t}(\bm{E}_{t-1}-\nabla_{\bm{E}}\mathcal{L}(\bm{E}_{t-1})/\tau_E^t)$ \\ \hline
\end{tabular}\label{Tab_sol_E}

\end{table}

To determine $\tau_E^t$, we estimate $L_E^t$ as $\hat{L}_E^t=\xi\Vert \bar{\bm{W}}_3^t\Vert_2/\sigma^2=\xi\Vert\bm{1}_d^\top\bm{W}_3^t\Vert_\infty/\sigma^2$ where $\xi$ is a  sufficiently large constant. Equivalently, we set $\tau_E^t=\xi\Vert\bm{1}_d^\top\bm{W}_3^t\Vert_\infty/\sigma^2$ where $\xi$ is a  sufficiently large constant.
The following lemma indicates that updating $\bm{E}$ by Table \ref{Tab_sol_E} is nonexpansive.
\begin{lemma}\label{lem_LE}
Let $\bm{E}_t$ be the solution of \eqref{Eq.prob_E1}, where $\tau_E^t=\xi\Vert\bm{1}_d^\top\bm{W}_3^t\Vert_\infty/\sigma^2$ and $\xi$ is sufficiently large. Then
\begin{equation*}
\begin{aligned}
&\mathcal{J}(\bm{D}_{t},\bm{C}_{t},\bm{E}_{t})-\mathcal{J}(\bm{D}_{t},\bm{C}_{t},\bm{E}_{t-1})\\
\leq &-\dfrac{\tau_E^t-L_E^t}{2}\Vert \bm{E}_t-\bm{E}_{t-1}\Vert_F^2\ \leq 0.
\end{aligned}
\end{equation*}
\end{lemma}
\begin{remark}
Empirically, we found that Lemma \ref{lem_LE} often holds when $\xi=1$. It means $\hat{L}_E^t$ well approximates the Lipschitz constant of $\nabla_{\bm{E}}\mathcal{L}(\bm{E})$ at iteration $t$.
\end{remark}

\subsection{The overall algorithm}
The optimization for RNLMF is summarized in Algorithm \ref{alg.RNLMF}. The hyper-parameter $\sigma$ controls the smoothness of Gaussian RBF kernel and provides us flexibility to handle nonlinearity of different levels. We set $\sigma=cn^{-2}\sum_{ij}\Vert\hat{\bm{x}}_i-\hat{\bm{x}}_j\Vert$, where $c$ is a constant such as $0.5$, $1$, or $3$. When the data have strong nonlinearity, we use a smaller $\sigma$ (smaller $c$); otherwise, we use a larger $\sigma$ (larger $c$).

In Algorithm \ref{alg.RNLMF}, when $\mathcal{R}(\bm{C})=\Vert\bm{C}\Vert_\ast$ or $\Vert\bm{C}\Vert_1$, the convergence with $\eta=0$ is guaranteed by Theorem \ref{Eq.converge}, although a nonzero $\eta$ (e.g., 0.5) works better in practice. When $\mathcal{R}(\bm{C})=\tfrac{1}{2}\Vert\bm{C}\Vert_F^2$, the convergence is similar and the proof is omitted for simplicity. Proving the algorithm converges to a critical point is out of the scope of this paper, though it may be accomplished by following the methods used in \cite{bolte2014proximal}.

\begin{theorem}\label{Eq.converge}
Let $\lbrace \bm{C}_t,\bm{D}_t,\bm{E}_t\rbrace$ be the sequence generated by Algorithm \ref{alg.RNLMF} with $\eta=0$. Suppose $\tau_D^t$, $\mu$, and $\xi$ are sufficiently large. Then
\begin{equation*}
\begin{aligned}
&\lim_{t\rightarrow\infty} \mathcal{J}(\bm{D}_t,\bm{C}_t,\bm{E}_t)-\mathcal{J}(\bm{D}_{t-1},\bm{C}_{t-1},\bm{E}_{t-1})=0,\\
&\lim_{t\rightarrow\infty}\Vert \bm{D}_t-\bm{D}_{t-1}\Vert_F=0,\\
&\lim_{t\rightarrow\infty}\Vert \bm{C}_t-\bm{C}_{t-1}\Vert_F=0,\\
&\lim_{t\rightarrow\infty}\Vert \bm{E}_t-\bm{E}_{t-1}\Vert_F=0.
\end{aligned}
\end{equation*}
\end{theorem}

In Section \ref{Sec.update_D}, when $\mathcal{R}(\bm{C})=\Vert\bm{C}\Vert_1$ or $\Vert\bm{C}\Vert_\ast$, the subproblem of $\bm{C}$ has no closed-form solution, which may slow down convergence. We found that using $\Vert\bm{C}\Vert_F^2$ in the early iterations and then switching to $\Vert\bm{C}\Vert_1$ or $\Vert\bm{C}\Vert_\ast$ can speed up convergence. Nevertheless, our numerical results in Section \ref{sec.syn} showed that $\Vert\bm{C}\Vert_F^2$ outperformed $\Vert\bm{C}\Vert_1$ and $\Vert\bm{C}\Vert_\ast$.

\section{Time and space complexity}
In Algorithm 1, we need to store $\hat{\bm{X}}\in\mathbb{R}^{m\times n}$, $\bm{E}\in\mathbb{R}^{m\times n}$, $\bm{D}\in\mathbb{R}^{m\times d}$, $\bm{C}\in\mathbb{R}^{d\times n}$, $\mathcal{K}(\hat{\bm{X}}-\bm{E},\bm{D})\in\mathbb{R}^{n\times d}$, and $\mathcal{K}(\bm{D},\bm{D})\in\mathbb{R}^{d\times d}$. Then the space complexity of RNLMF is as $O(mn+md+dn+d^2)$ or $O(mn+dn)$ equivalently because of $m,d<n$. In each iteration of Algorithm 1, the main computational cost is from the computation of  $\mathcal{K}(\hat{\bm{X}}-\bm{E},\bm{D})$, the inverse of a $d\times d$ matrix (or the SVD of a $d\times n$ matrix) in Table \ref{Tab_sol_C}, the inverse of a $d\times d$ matrix in updating $\bm{D}$, and the related multiplications in \eqref{Eq.C_sol_1}, \eqref{Eq.D_sol_1}, and \eqref{Eq.gradE}. Then the time complexity in each iteration of RNLMF is $O(d^3+d^2n+d^2m+dmn)$.

The time and space complexities of RPCA \cite{RPCA}, LRR \cite{LRR_PAMI_2013}, NLRR \cite{pmlr-v51-shen16}, SSC \cite{SSC_PAMIN_2013}, RDL (problem \eqref{Eq.RCDL_3}), and RNLMF are compared in Table \ref{Tab.cc}, where truncated (top-$r$) SVD is considered in LRR and $\mathcal{R}(\bm{C})=\Vert\bm{C}\Vert_1$ or $\Vert\bm{C}\Vert_F^2$ are considered for RNLMF. We see that when $n$ is large, SSC and LRR have high time and space complexities. The computational costs of RNLMF and RDL are similar, though in real applications the $d$ in RNLMF should be larger than that in RDL. But RNLMF is able to handle data generated by more complex models.

\begin{table}[h]
\centering
\begin{threeparttable}
\caption{Time and space complexities}\label{Tab.cc}
\begin{tabular}{|l|l|l|}
\hline
	&  Time complexity & Space complexity \\
\hline
RPCA & $O(m^2n)$	&  $O(mn)$  \\
LRR & $O(mn^2+rn^2)$	&  $O(mn+n^2)$  \\
NLRR & $O(m^2n+rmn)$	&  $O(mn+rn)$  \\
SSC & $O(mn^2)$	&  $O(mn+n^2)$  \\
RDL & $O(d^2n+dmn)$	&  $O(mn+dn)$  \\
RNLMF &  $O(d^2n+dmn)$ & $O(mn+dn)$ \\
\hline
\end{tabular}
*Assume  $r<m<n$, $r<d<n$, and $d^2\leq mn$.
\end{threeparttable}
\end{table}

\section{Out-Of-Sample Extension of RNLMF}

It is worth mentioning that in an online fashion, the dictionary $\bm{D}$ given by Algorithm \ref{alg.RNLMF} can be used to denoise a new data matrix $\hat{\bm{X}}'$ generated from the same model as $\hat{\bm{X}}$. The approach is shown in Algorithm \ref{alg.RNLMF_ose}.

\begin{algorithm}[h!]
\caption{Out-of-sample extension of RNLMF}
\label{alg.RNLMF_ose}
\begin{algorithmic}[1]
\Require
$\hat{\bm{X}}'$, $\bm{D}$ (given by Algorithm \ref{alg.RNLMF}), $t_{\text{iter}}$.
\State Initialize: $\bm{E}'=\bm{0}$, $\bm{C}'=\bm{0}$, $t=0$.
\Repeat
\State $t\leftarrow t+1$.
\State Update $\bm{C}'$ using Table \ref{Tab_sol_C}.
\State Update $\bm{E}'$ using Table \ref{Tab_sol_E}.
\Until{converged or $t=t_{\text{iter}}$}
\Ensure $\bm{X}'=\hat{\bm{X}}'-\bm{E}'$, $\bm{C}'$.
\end{algorithmic}
\end{algorithm}

\section{Subspace clustering by RNLMF}
RNLMF can be regarded as a robust nonlinear feature extraction method, thus the feature matrix $\bm{C}$ can be used for clustering. One may perform SSC \cite{SSC_PAMIN_2013} or LRR \cite{LRR_PAMI_2013} on $\bm{C}$, which, however, is not efficient. We propose to compute an affinity matrix by solving the following least squares problem
\begin{equation}
\mathop{\textup{minimize}}_{\bm{A}}\tfrac{1}{2}\Vert \bm{C}-\bm{C}\bm{A}\Vert_F^2+\tfrac{\gamma}{2}\Vert\bm{A}\Vert_F^2,
\end{equation}
where $\gamma$ is a penalty parameter.
The solution is $\bm{A}=(\bm{C}^\top\bm{C}+\gamma\bm{I})^{-1}\bm{C}^\top\bm{C}$. Note that least squares regression is also effective in subspace segmentation \cite{LSRSCLu2012}.  Then let $\bm{A}\leftarrow\vert\bm{A}\vert$, set $\textup{diag}(\bm{A})=\bm{0}$, and keep the largest $\kappa$ entries of each column of $\bm{A}$ and discard the other entries. A normalization is performed on each column of $\bm{A}$: $\bm{a}_j=\bm{a}_j/\max(\bm{a}_j)$, $j=1,2,\ldots,n$. To ensure a symmetric affinity matrix, we set $\bm{A} \leftarrow (\bm{A}+\bm{A}^\top)/2$. Finally, spectral clustering is performed on $\bm{A}$. The procedures are summarized in Algorithm \ref{alg.RNLMF_sc}. The role of Procedures $3\sim5$ is similar to the post-processing in SSC and LRR, making the affinity matrix more compact. It is worth mentioning that using the Woodbury identity,  line 2 in Algorithm \ref{alg.RNLMF_sc} is equivalent to $\bm{A}=\vert\gamma^{-1}\bm{C}^\top(\bm{I}+\gamma^{-1}\bm{C}\bm{C}^\top)^{-1}\bm{C}\vert$, which reduced the computational cost.

\begin{algorithm}[h!]
\caption{Subspace clustering by RNLMF}
\label{alg.RNLMF_sc}
\begin{algorithmic}[1]
\Require
$\hat{\bm{X}}$, $k$, $\kappa$, $\gamma$.
\State Compute $\bm{C}$ using Algorithm \ref{alg.RNLMF}.
\State $\bm{A}=\vert(\bm{C}^\top\bm{C}+\gamma\bm{I})^{-1}\bm{C}^\top\bm{C}\vert$ and set $\textup{diag}(\bm{A})=\bm{0}$.
\State Keep only the largest $\kappa$ entries of each column of $\bm{A}$.
\State For $j=1,2,\ldots,n$, $\bm{a}_j\leftarrow\bm{a}_j/\max(\bm{a}_j)$.
\State $\bm{A}\leftarrow(\bm{A}+\bm{A}^\top)/2$.
\State Perform spectral clustering on $\bm{A}$ with cluster number $k$.
\Ensure $k$ clusters of $\hat{\bm{X}}$.
\end{algorithmic}
\end{algorithm}

When the number of data points is very large (e.g. $n>10^5$), we cannot use Algorithm \ref{alg.RNLMF_sc} to cluster the whole dataset because the high space cost of $\bm{A}$. Recently a few large-scale subspace clustering methods have been proposed \cite{cai2014large,you2018scalable} and they often take advantage of exemplars or landmark points selection to cluster large-scale datasets but cannot effectively handle sparse noise. Similar ideas may apply to our RNLMF, which however is out of the scope of this paper.

\vspace{-5pt}
\section{Experiments on Synthetic Data}\label{sec.syn}
We generate synthetic data by
$$\bm{x}=f(\bm{z}), f\in\lbrace{F^{1},F^{2},\ldots,F^{k}}\rbrace,$$
where each $F^{j}:\mathbb{R}^3\rightarrow\mathbb{R}^{30}$, $1\leq j\leq k$ is an order-$3$ polynomial mapping and $\bm{z}=[z_1,z_2,z_3]^\top\sim\mathcal{U}(-1,1)$.
The model can be reformulated as
$$\bm{x}=\bm{P}\tilde{\bm{z}}, \bm{P}\in\lbrace{\bm{\Gamma}^{1},\bm{\Gamma}^{2},\ldots,\bm{\Gamma}^{k}}\rbrace,$$
where $\bm{\Gamma}^j\in\mathbb{R}^{30\times 19}\sim\mathcal{N}(0,1)$ for $1\leq j\leq k$, and $\tilde{\bm{z}}\in\mathbb{R}^{19}$ consists of order-1, 2 and 3 polynomial features of $\bm{z}$. For each fixed $\bm{\Gamma}^j$, we generate $300$ random samples of $\bm{x}$. Then we obtain a matrix $\bm{X}\in\mathbb{R}^{30\times 300k}$, which is full-rank when $k\geq 2$.
We then add sparse noise to $\bm{X}$, i.e.
$\hat{\bm{X}}=\bm{X}+\bm{E}$, where $\tfrac{1}{9000k}\sum_{ij}\mathbb{1}(E_{ij}\neq 0)=\rho$ and the nonzero entries of $\bm{E}$ are drawn from $\mathcal{N}(0,\sigma_e^2)$. The locations of nonzero entries are chosen uniformly at random by sampling without replacement. Denote the standard deviation of the entries of $\bm{X}$ by $\sigma_x$.

The denoising performance is evaluated by the normalized root-mean-square-error:
\begin{equation*}
\textup{RMSE}:=\Vert \bm{X}-\check{\bm{X}}\Vert_F/\Vert \bm{X}\Vert_F,
\end{equation*}
where $\check{\bm{X}}$ denotes the recovered matrix. All results we report in this paper are the average of 20 repeated trials. In RNLMF, we set $d=2mk$, choose $\lambda_C$ from $[1,5,10]/10^3$, and choose $\lambda_E$ from $[0.3,0.5,1,2]/10^3$; we set $\sigma=n^{-2}\sum_{ij}\Vert\hat{\bm{x}}_i-\hat{\bm{x}}_j\Vert$.  Such parameter settings are utilized throughout this paper, unless stated otherwise.

\subsection{Choice of $\mathcal{R}(\bm{C})$}
\begin{figure}[h]
\centering
\includegraphics[width=7cm,trim={32 0 35 0},clip]{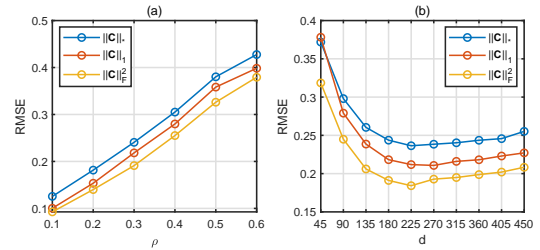}

\caption{RNLMF with different $\mathcal{R}(\bm{C})$: (a) $k=3$, $\sigma_e/\sigma_x=1$, $d=2mk$, and different $\rho$; (b) $k=3$, $\sigma_e/\sigma_x=1$, $\rho=0.3$, and different $d$.}
\label{fig_syn_Rcompare}

\end{figure}

Figure \ref{fig_syn_Rcompare}(a) shows the RMSE of RNLMF with different penalty operators of $\bm{C}$ when $k=3$ and varying $\rho$, for which we have sufficiently tuned $\lambda_C$ and $\lambda_E$ separately for each regularizer  $\mathcal{R}(\bm{C})$. We see that $\Vert\bm{C}\Vert_F^2$ always outperform $\Vert\bm{C}\Vert_1$ and $\Vert\bm{C}\Vert_\ast$. In Figure \ref{fig_syn_Rcompare}(b), where $k=3$ and $\rho=0.3$, $\Vert\bm{C}\Vert_F^2$ outperformed $\Vert\bm{C}\Vert_1$ and $\Vert\bm{C}\Vert_\ast$, in all choices of $d$ (the number of columns of $\bm{D}$). In addition, RNLMF is relatively not sensitive to $d$ provided that $d$ is large enough (e.g. $d\geq 135$). The advantage of $\Vert\bm{C}\Vert_F^2$ over $\Vert\bm{C}\Vert_1$ and $\Vert\bm{C}\Vert_\ast$ may result from: (1) $\Vert\bm{C}\Vert_F^2$ leads to a closed-form solution for updating $\bm{C}$, which enables the optimization of RNLMF to obtain a better stationary point; (b) $\phi(\bm{X})$ is low-rank such that the denoising problem does not benefit from enforcing $\bm{C}$ to be sparse; (c) enforcing $\bm{C}$ to be low-rank reduces the compactness of $\phi(\bm{D})$. In the remaining of this paper, we only use $\mathcal{R}(\bm{C})=\Vert\bm{C}\Vert_F^2$ in RNLMF.

\subsection{Influence of hyper-parameters in RNLMF}
\begin{figure}[h]
\centering
\includegraphics[width=6cm, trim={0 0 0 15},clip]{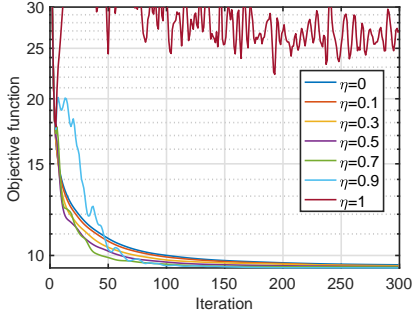}
\caption{Influence of $\eta$ in the optimization of RNLMF}
\label{fig_syn_eta}
\vspace{-10pt}
\end{figure}

Figure \ref{fig_syn_eta} shows the influence of $\eta$ in the optimization of RNLMF with $\mathcal{R}(\bm{C})=\Vert\bm{C}\Vert_F^2$, in the case of $k=3$ and $\rho=0.3$. We see that when $\eta$ increases, the objective function converges faster. But when $\eta$ is too large, the algorithm may diverge.

\begin{figure}[h]
\begin{subfigure}[h]{4cm}
\includegraphics[width=4.5cm, trim={0 0 0 -5},clip]{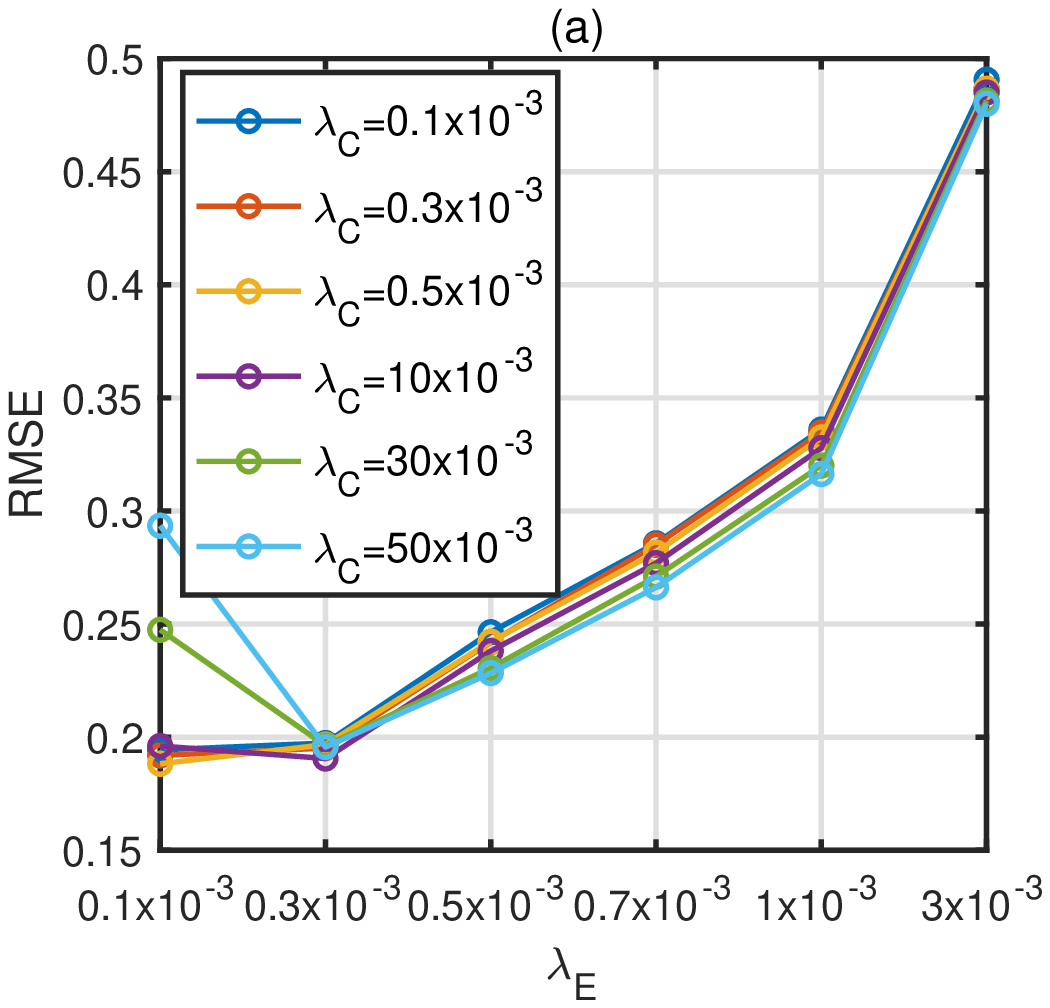}
\end{subfigure}
\hspace{10pt}
\begin{subfigure}[h]{4cm}
\includegraphics[width=4.4cm, trim={0 0 0 5},clip]{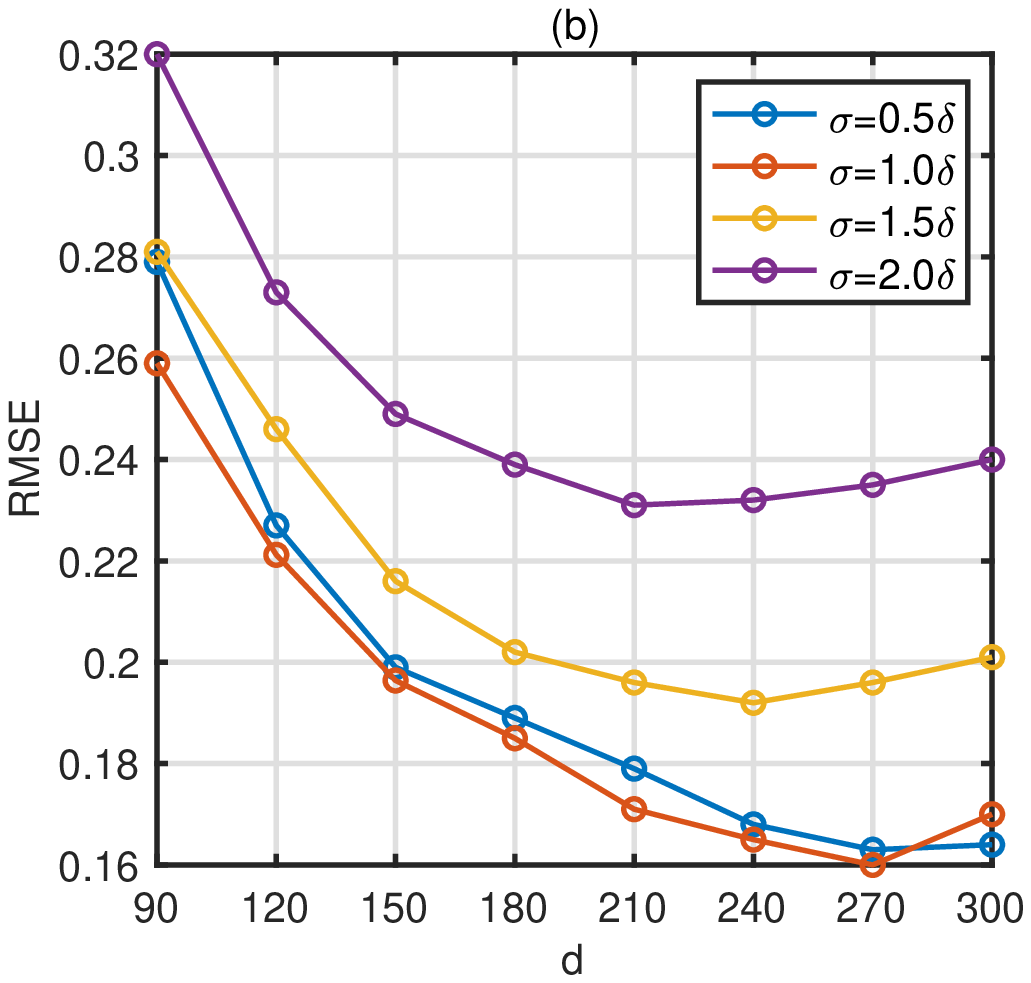}
\end{subfigure}
\caption{Influence of $\lambda_C$, $\lambda_E$, $\sigma$, and $d$ in RNLMF ($k=3$, $\rho=0.3$, $\delta=n^{-2}\sum_{ij}\Vert\hat{\bm{x}}_i-\hat{\bm{x}}_j\Vert$).}
\label{fig_syn_par}
\vspace{-10pt}
\end{figure}

Figure \ref{fig_syn_par}(a) shows the sensitivity of our method to the hyper-parameters $\lambda_C$ and $\lambda_E$ on synthetic data ($k=3$, $\rho$=0.3). We see that our RNLMF is not sensitive to $\lambda_C$ and has low recovery error when $0.1\times 10^{-3}\leq\lambda_E\leq 0.7\times 10^{-3}$. Figure \ref{fig_syn_par}(b) shows the influence of the hyper-parameter $\sigma$ of Gaussian RBF kernel and $d$ in RNLMF. We see that $\sigma=0.5\delta$ and $1.0\delta$ outperformed $\sigma=1.5\delta$ and $2\delta$. In addition, $d=270$ is the best to $\sigma=0.5\delta$ and $1.0\delta$, $d=240$ is the best to $\sigma=1.5\delta$, and $d=210$ is the best to $\sigma=2\delta$. It indicates that when $\sigma$ is small, the optimal $d$ should be large.

\subsection{Comparison with RPCA, LRR, SSC, and RDL in denoising}
Since the kernel methods \cite{van2013design,7299018,liu2015robust,quan2016equiangular,fan2019online} do not provide denoised matrix $\bm{X}$, we compare RNLMF with RPCA (problem \eqref{Eq.RPCA}, solved by ADMM), RDL (problem \eqref{Eq.RCDL_3}, solved by PALM \cite{bolte2014proximal}),  LRR \cite{LRR_PAMI_2013}, and SSC \cite{SSC_PAMIN_2013}. First, we consider sparse noise and  use $\mathcal{R}(\bm{E})=\Vert\bm{E}\Vert_1$ for all compared methods. In RPCA, the parameter $\lambda$ is chosen from $[0.5,0.75,1,1.5,2,2.5,3]/\sqrt{n}$. In RDL, we set the number of dictionary atoms as $0.5mk$ or $mk$, choose $\lambda_C$ from $[1,3,5,10]/10^3$, and choose $\lambda_E$ from $[0.03,0.05,0.07,0.1,0.15,0.2]$. The parameters in LRR and SSC are carefully tuned to provide the best denoising performance as possible. The parameter setting in RNLMF has been stated in the beginning of Section \ref{sec.syn}.

\begin{figure}[h!]
\centering
\includegraphics[width=7.5cm,trim={30 15 35 20},clip]{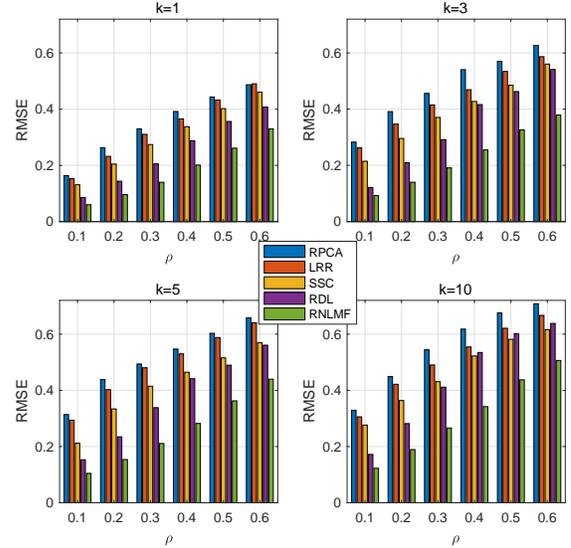}
\vspace{-5pt}
\caption{RMSE on synthetic data ($\tfrac{\sigma_e}{\sigma_x}=1$, different $k$ and $\rho$).}
\label{fig_syn_1}
\end{figure}

Figure \ref{fig_syn_1} shows the recovery errors when $\sigma_e/\sigma_x=1$ and $\rho$ and $k$ vary. When $k$ or $\rho$ increase, the recovery task becomes more difficult. SSC and RDL outperformed RPCA and LRR. The reason is that sparse representation is more effective than low-rank model in handling high-rank matrices. In every case of Figure \ref{fig_syn_1}, the RMSE of our RNLMF is much lower than those of other methods. The improvement given by RNLMF is owing to the ability of RNLMF to handle nonlinear data and full-rank matrices, which are challenges for other methods.

\begin{figure}[h!]
\centering
\includegraphics[width=4.5cm,trim={5 0 20 15},clip]{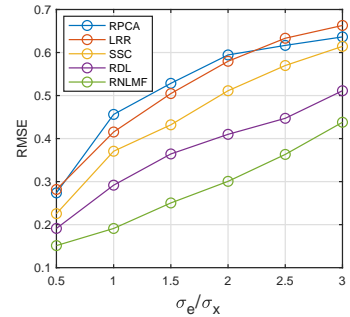}

\caption{RMSE on synthetic data ($k=3$, $\rho=0.3$).}
\label{fig_syn_2}

\end{figure}

We investigate the influence of the noise magnitude on the performance of five methods in the case of $k=3$ and $\rho=0.3$, shown in Figure \ref{fig_syn_2}. We see that RNLMF consistently outperforms other methods with different $\sigma_e/\sigma_x$.

\begin{figure}[h!]
\centering
\includegraphics[width=7.5cm,trim={50 0 60 5},clip]{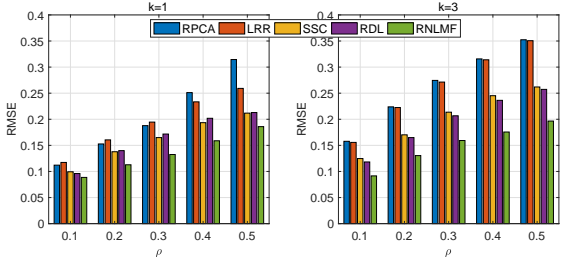}

\caption{RMSE on synthetic data with column-wise noise.}
\label{fig_syn_3}

\end{figure}

To evaluate the ability of all methods to handle column-wise noise, we add independent and identically distributed noise drawn from $\mathcal{N}(0,\sigma_x^2)$ to a fraction (denoted by $\rho$) of columns of $\bm{X}$. Therefore, we use $\mathcal{R}(\bm{E})=\Vert\bm{E}\Vert_{2,1}$ in all compared methods. Shown in Figure \ref{fig_syn_3}, the RMSE of RNLMF is the lowest in every case.

\section{Experiments on Image Data}
To test the performance of our method on real data, we consider the following four image datasets.
\begin{itemize}
\item COIL20\cite{coil20}/COIL100\cite{coil100}, images of 20/100 objects. Each object has 72 images of different
poses.
\item Extended Yale Face database B (Yale Face for short) \cite{Dataset_ExtendYaleB}, face images of 38 subjects. Each subject has about 64 images under various illumination conditions.
\item AR Face database (a subset) \cite{ARfacedata}, consisting of the face images of 50 males and 50 females \cite{4483511}. Each subject has 26 images with different facial expressions, illumination conditions, and occlusions.
\end{itemize}
We resize the images in AR Face to $33\times 24$ and resize the images in the other three databases to $20\times 20$.
We consider two cases of image corruption. In the first case, we add salt-and-pepper noise of density 0.25 to $30\%$ of the images. In the other case, for each data set, we occlude $30\%$ of the images with a block mask of size $0.25h\times 0.25w$ and position random, where $h$ and $w$ are the height and length of the images. Since the two cases are sparse noise patterns, we use $\mathcal{R}(\bm{E})=\Vert\bm{E} \Vert_1$ in RPCA, LRR, SSC, RDL, and RNLMF.
For COIL20, Yale Face, AR Face, and COIL100, the $d$ in RDL is set to $196$, $196$, $256$, and $512$ respectively, while the $d$ in RNLMF is set to $256$, $256$, $512$, and $768$ respectively.

\begin{figure}[h!]
\centering
\includegraphics[width=8cm,trim={00 00 00 00},clip]{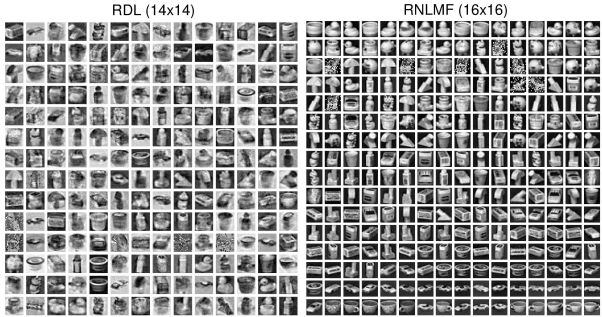}
\caption{The dictionaries learned from COIL20.}
\label{fig_dict_1}
\vspace{-10pt}
\end{figure}

\begin{figure}[h!]
\centering
\includegraphics[width=8cm,trim={0 0 0 0},clip]{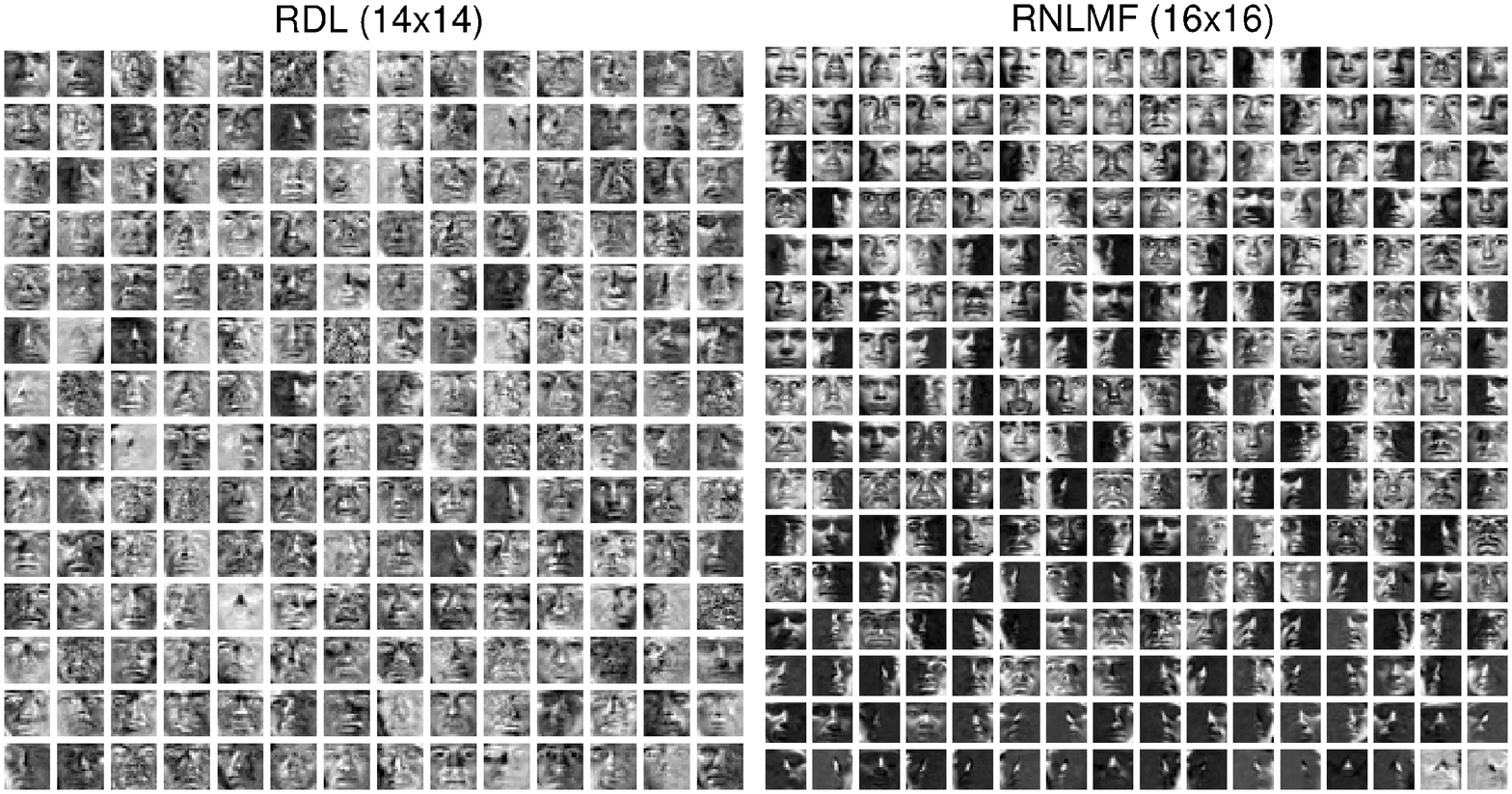}
\caption{The dictionaries learned from Yale Face.}
\label{fig_dict_2}
\end{figure}

Compared to Yale Face and AR Face, the nonlinearity of data structure in COIL20 and COIL100 are much higher because of the different posses of the objects. For each data set, we stack the pixels of each image as a matrix column and then form a matrix $\bm{X}$ of size $m\times n$, where $m=hw$ and $n$ is the number of images. The metric $R:=\Vert\bm{X}\Vert_\ast/\Vert\bm{X}\Vert_F$ can be utilized to compare the rank of the data matrices of the four data sets. For COIL20, COIL100, Yale Face, and AR Face, the values of $R$ are 5.17, 5.03, 4.65, and 4.33 respectively. Thus for COIL20 and COIL100, we set $\sigma=n^{-2}\sum_{ij}\Vert\hat{\bm{x}}_i-\hat{\bm{x}}_j\Vert$ in RNLMF; for Yale Face and AR Face, we set $\sigma=3n^{-2}\sum_{ij}\Vert\hat{\bm{x}}_i-\hat{\bm{x}}_j\Vert$ and $5n^{-2}\sum_{ij}\Vert\hat{\bm{x}}_i-\hat{\bm{x}}_j\Vert$ in RNLMF respectively. We expect that the improvement given by RNLMF on COIL20 and COIL100 are higher than those on Yale Face and AR Face.

\begin{table*}[h!]
\centering
\small
\caption{RMSE ($\%$) of denoising on the noisy image data}
\begin{tabular}{|c|c|c|c|c|c|c|}
\hline
Data	&{\footnotesize Noise}            & RPCA      & LRR 	&SSC 	& RDL        & RNLMF        \\ \hline
\multirow{3}{*}{{\footnotesize COIL20}}		&Random       & 12.28$\pm$0.04 &	13.84$\pm$0.09	&12.05$\pm$0.08	& 11.04$\pm$0.14 & \textbf{9.31}$\pm$0.44  \\
		&Occlusion        & 20.54$\pm$0.31 &	18.34$\pm$0.45	&15.26$\pm$0.38	& 14.89$\pm$0.63 & \textbf{10.25}$\pm$0.59 \\
		&{\footnotesize Random+Occlusion} & 21.26$\pm$0.23 &	21.81$\pm$0.24	& 18.84$\pm$0.23	& 18.91$\pm$0.39  & \textbf{12.67}$\pm$0.56 \\ \hline
\multirow{3}{*}{{\footnotesize COIL100}}		&Random       & 13.75$\pm$0.02 &	13.28$\pm$0.05	& 12.39$\pm$0.06	& 11.89$\pm$0.09 & \textbf{9.15}$\pm$0.03  \\
		&Occlusion        & 20.98$\pm$0.05 & 18.95$\pm$0.11		&18.09$\pm$0.12	 & 17.41$\pm$0.15 & \textbf{13.82}$\pm$0.17 \\
		&{\footnotesize Random$\&$Occlusion} & 23.93$\pm$0.22 & 21.26$\pm$0.18		& 20.05$\pm$0.25	& 20.76$\pm$0.58 & \textbf{14.86}$\pm$0.32 \\ \hline
\multirow{3}{*}{{\footnotesize Yale Face}}		&Random       & 9.25$\pm$0.03 &	12.87$\pm$0.03	& 13.04$\pm$0.04	& 9.71$\pm$0.11 & \textbf{7.71}$\pm$0.14  \\
		&Occlusion       & 15.80$\pm$0.15 &	13.41$\pm$0.14	& 13.28$\pm$0.11	& 13.68$\pm$0.19 & \textbf{10.96}$\pm$0.14 \\
		&{\footnotesize Random+Occlusion} & 17.07$\pm$0.14 &	16.05$\pm$0.13	&	16.66$\pm$0.12 & 16.24$\pm$0.28  & \textbf{12.20}$\pm$0.29 \\ \hline
\multirow{3}{*}{{\footnotesize AR Face}}		&Random       & 8.41$\pm$0.02 &	9.26$\pm$0.03	& 9.15$\pm$0.02	& 7.81$\pm$0.11 & \textbf{6.01}$\pm$0.02  \\
		&Occlusion        & 13.25$\pm$0.04 & 12.19$\pm$0.13		&11.57$\pm$0.11	& 11.89$\pm$0.16 & \textbf{10.65}$\pm$0.07 \\
		&{\footnotesize Random+Occlusion} & 13.49$\pm$0.05 &	13.18$\pm$0.07	& 13.04$\pm$0.06	& 12.71$\pm$0.12  & \textbf{11.75}$\pm$0.13 \\ \hline
\end{tabular}\label{Tab_real_dn}
\end{table*}

\subsection{Denoising result}

The dictionaries given by RDL and RNLMF on COIL20 and Yale Face are visualized in Figure \ref{fig_dict_1} and Figure \ref{fig_dict_2}. We see that the dictionary of RNLMF consists of real images, because Gaussian RBF kernel in RNLMF plays a role of smooth interpolation and RNLMF constructs a set of ``landmark" points to represent all data point as accurate as possible.

\begin{figure}[h!]
\centering
\includegraphics[width=7.5cm,trim={0 80 35 20},clip]{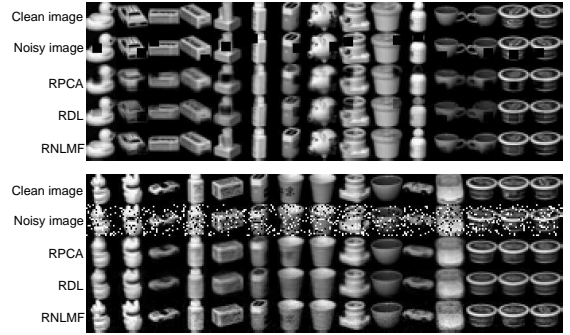}
\caption{Denoising COIL20}
\label{fig_imr_1}
\end{figure}

\begin{figure}[h!]
\centering
\includegraphics[width=7.5cm,trim={0 80 35 20},clip]{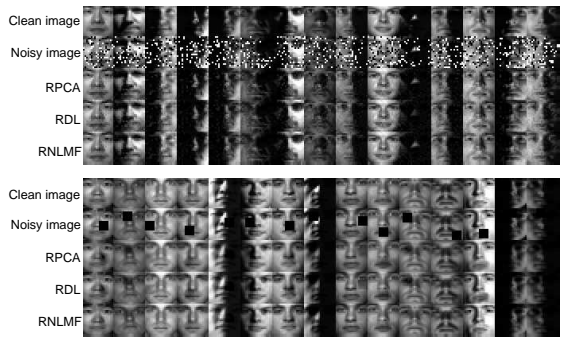}
\caption{Denoising Yale Face}
\label{fig_imr_2}
\end{figure}

Figure \ref{fig_imr_1} and Figure \ref{fig_imr_2} show some examples of the original images, noisy images, and recovered images of COIL20 and Yale Face. The denoising performance of RNLMF is better than those of RPCA and RDL. The average RMSE and its standard deviation of 20 repeated trials are reported in Table \ref{Tab_real_dn}. Note that we actually performed NLRR \cite{pmlr-v51-shen16} rather than LRR \cite{LRR_PAMI_2013} on COIL100 because the data size is large, though we still use the name LRR for consistency. In the table, RNLMF outperformed other methods significantly in all cases.

\vspace{-10pt}

\begin{table}[h!]
\centering
\caption{Clustering error ($\%$) on the original image data}
\begin{tabular}{|p{1cm}|p{0.55cm}|p{0.55cm}|p{0.55cm}|p{0.65cm}|p{0.65cm}|p{0.6cm}|p{0.75cm}|}\hline
  &\makecell{\hspace{-3pt}\ \ LRR\ }    & \makecell{\hspace{-3pt}\ \ SSC\ } &\makecell{\hspace{-2pt}KSSC}  &\makecell{\hspace{-5pt} GMC-\\\hspace{-3pt}LRSSC}& \makecell{\hspace{-5pt}$S_0/\ell_0$-\\\hspace{-3pt}LRSSC} &\makecell{EKSS}&\makecell {\hspace{-3pt}RNLMF} \\ \hline
\multirow{1}{*}{COIL20}& 25.21 & 14.36 &21.94 &25.83 &18.40 &13.47 & \textbf{13.13 }\\
\multirow{1}{*}{COIL100}& 52.28  & 44.63  &44.67&55.04& 46.99  &28.57& \textbf{23.53} \\
\multirow{1}{*}{Yale Face}& 13.26  & 21.75 & 20.55 &23.20 &12.01&14.31  &\textbf{10.77}  \\
\multirow{1}{*}{AR Face}& 19.27  & 24.61 &25.54&18.92&  27.15& 22.65  & \textbf{13.88}\\ \hline
\end{tabular}\label{Tab_real_cluster_org}
\end{table}

\begin{table}[h!]
\centering
\caption{Clustering error ($\%$) on the noisy image data}
\begin{tabular}{|p{1cm}|p{0.55cm}|p{0.55cm}|p{0.55cm}|p{0.65cm}|p{0.65cm}|p{0.6cm}|p{0.75cm}|}\hline
  &\makecell{\hspace{-3pt}\ \ LRR\ }    & \makecell{\hspace{-3pt}\ \ SSC\ } &\makecell{\hspace{-2pt}KSSC}  &\makecell{\hspace{-5pt} GMC-\\\hspace{-3pt}LRSSC}& \makecell{\hspace{-5pt}$S_0/\ell_0$-\\\hspace{-3pt}LRSSC} &\makecell{EKSS}&\makecell {\hspace{-3pt}RNLMF} \\ \hline
\multirow{1}{*}{COIL20}& 34.79 & 24.65 &40.42 &31.11&28.68&21.11 & \textbf{14.03} \\
\multirow{1}{*}{COIL100}& 65.54  & 52.17  &61.56 & 65.97 & 49.58 &54.17 & \textbf{34.18} \\
\multirow{1}{*}{Yale Face}& 55.82  & 35.21& 64.25 &33.43 &33.14 &\textbf{18.97}&21.13  \\
\multirow{1}{*}{AR Face}& 63.38  & 39.88 &61.50  &36.65 &38.92   & 28.96 & \textbf{23.27}\\ \hline
\end{tabular}\label{Tab_real_cluster_noisy}
\end{table}

\subsection{Clustering result}
We check the clustering performance of RNLMF compared with LRR \cite{LRR_PAMI_2013} \cite{pmlr-v51-shen16}, SSC \cite{SSC_PAMIN_2013}, KSSC \cite{KSSC}, GMC-LRSSC \cite{8573150}, $S_0/\ell_0$-LRSSC \cite{8573150}, and EKSS \cite{lipor2017subspace} on the four datasets. Since KSSC, GMC-LRSSC, $S_0/\ell_0$-LRSSC, and EKSS cannot handle sparse noise we first process the data by RPCA and then implement the four clustering methods. Moreover, in line with \cite{lipor2017subspace}, we perform EKSS on the features extracted by PCA rather than the pixel values; otherwise, the clustering error of EKSS is too large. In Algorithm \ref{alg.RNLMF_sc}, we set $\gamma=0.01$;  we set $\kappa=5$ on Yale Face and COIL100, and set $\kappa=15$ on the COIL20 and AR Face.  The hyper-parameters of other methods are carefully tuned to provide their best performances. The clustering errors on the original data are reported in Table \ref{Tab_real_cluster_org}, in which RNLMF has the lowest clustering error in every case. As shown in Table \ref{Tab_real_cluster_noisy}, RNLMF outperformed other methods significantly on noisy COIL20, COIL100 and AR Face. The main reason is that RNLMF is more effective than other methods in handling high-rank matrices corrupted by sparse noise. EKSS benefits a lot from the preprocessing of RPCA especially on Yale Face, of which the matrix rank is much lower than those of COIL20 and COIL100.

\begin{table*}[h]
\centering
\caption{RMSE ($\%$) and MAE ($\%$) on motion capture data}
\begin{tabular}{|c|c|c|c|c|c|c|c|}\hline
&subject & $\rho$    & RPCA   &LRR &SSC & RDL  & RNLMF   \\ \hline	\hline
\multirow{4}{*}{RMSE}&\multirow{2}{*}{$\#$01}&0.1  & 11.19$\pm$0.10 &9.89$\pm$0.08	& 10.36$\pm$0.11 & 12.75$\pm$0.98  & \textbf{9.78}$\pm$0.45  \\
&					&0.3  &21.98$\pm$0.13 &17.54$\pm$0.10	& 18.66$\pm$0.12 & 23.51$\pm$1.22  & \textbf{11.82}$\pm$1.13 \\ \cline{2-8}
&\multirow{2}{*}{$\#$56}&0.1  & 8.67$\pm$0.11 &	9.06$\pm$ 0.08& 9.95$\pm$0.07 & 9.19$\pm$0.50  & \textbf{6.11}$\pm$0.33  \\
&					&0.3  &22.93$\pm$0.69 &17.73$\pm$0.07	& 18.50$\pm$0.07 & 19.03$\pm$0.99  & \textbf{11.27}$\pm$1.17 \\ \hline	 \hline
\multirow{4}{*}{MAE}&\multirow{2}{*}{$\#$01}&0.1  & 7.10$\pm$0.04 &7.38$\pm$0.08	&8.52$\pm$0.06   & 7.33$\pm$0.72  & \textbf{4.90}$\pm$0.40 \\
&					&0.3  &21.85$\pm$0.20 &22.31$\pm$0.14	& 23.75$\pm$0.09 & 23.61$\pm$1.76  & \textbf{10.11}$\pm$0.79 \\ \cline{2-8}
&\multirow{2}{*}{$\#$56}&0.1  & 5.68$\pm$0.33 & 6.57$\pm$0.05	& 8.23$\pm$0.05  & 5.68$\pm$0.28  & \textbf{4.11}$\pm$0.28  \\
&					&0.3  &22.23$\pm$0.40 &18.67$\pm$0.13	&22.48$\pm$0.09   & 17.89$\pm$0.48  & \textbf{11.04}$\pm$0.71 \\ \hline
\end{tabular}\label{Tab_real_mocap}
\end{table*}


\begin{table*}[h!]
\centering
\caption{RMSE ($\%$) and MAE ($\%$) on motion capture data (out-of-sample-extension)}
\begin{tabular}{|c|c|c|c|c|c|c|c|c|}\hline
& & & \multicolumn{3}{c|}{Training data} & \multicolumn{3}{c|}{Testing data}\\ \cline{2-9}
& subject & $\rho$    & RPCA & RDL& RNLMF   & RPCA & RDL& RNLMF\\ \hline	\hline
\multirow{4}{*}{RMSE} &\multirow{2}{*}{$\#$01}&0.1 &11.34$\pm$0.39	 	&12.20$\pm$1.19 & \textbf{9.21}$\pm$1.02 &13.78$\pm$0.33 &11.98$\pm$0.51 & \textbf{7.51}$\pm$0.87 \\
&						&0.3   &21.96$\pm$0.17 &23.06$\pm$1.14	& \textbf{11.45}$\pm$0.73 &20.35$\pm$0.34	 &22.83$\pm$0.96  & \textbf{10.86}$\pm$1.09\\  \cline{2-9}
&\multirow{2}{*}{$\#$56}&0.1   &8.65$\pm$0.11	&9.07$\pm$0.83 & \textbf{6.04}$\pm$0.52 &10.24$\pm$0.19	&8.92$\pm$1.04  & \textbf{5.57}$\pm$0.28 \\
&					&0.3  &22.51$\pm$1.14	 &19.87$\pm$1.19 & \textbf{10.89}$\pm$0.69 &17.37$\pm$0.50	  &18.94$\pm$1.16  &\textbf{10.58}$\pm$0.62\\ \hline	\hline
\multirow{4}{*}{MAE} & \multirow{2}{*}{$\#$01}&0.1 &	7.13$\pm$0.10	&7.37$\pm$0.58 & \textbf{4.83}$\pm$0.68 &9.65$\pm$0.13	 &7.96$\pm$0.41  & \textbf{4.93}$\pm$0.60 \\
&					&0.3  &22.03$\pm$0.32	&22.04$\pm$1.33 & \textbf{10.11}$\pm$0.32 &23.51$\pm$0.28	  &22.52$\pm$1.03  & \textbf{10.33}$\pm$0.28\\ 	\cline{2-9}
&\multirow{2}{*}{$\#$56}&0.1  &5.70$\pm$0.08	&5.81$\pm$0.42 & \textbf{4.07}$\pm$0.25 &7.86$\pm$0.90	 &5.93$\pm$0.36   & \textbf{4.05}$\pm$0.22 \\
&					&0.3   &22.04$\pm$0.69	&18.06$\pm$0.62 & \textbf{10.78}$\pm$0.53 &19.07$\pm$0.77	 &16.94$\pm$0.55  &\textbf{10.69}$\pm$0.50\\ \hline
\end{tabular}\label{Tab_real_mocap_ose}
\end{table*}

\section{Experiments on Motion Capture Data}
\begin{figure}
\centering
\includegraphics[width=6.5cm]{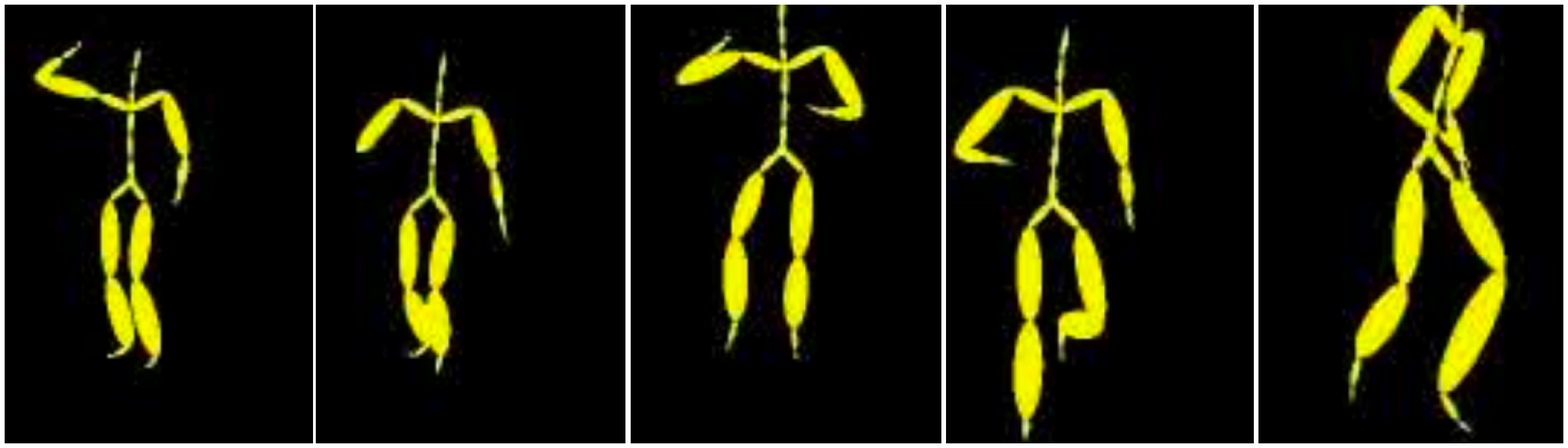}
\caption{A few examples of CMU motion capture data.}
\label{Fig.mocap}
\vspace{-10pt}
\end{figure}

Besides image data sets that consist of the images of multiple objects or subjects, many other data sets in computer vision as well as other areas can also form high-rank matrices. For example, in CMU motion capture database (http://mocap.cs.cmu.edu/), many subsets consist of the time-series trajectories of multiple human motions such as walking, jumping, stretching, and climbing; the dimension of the signal (the number of sensors) is 62, much smaller than the number of samples; the formed matrices are often high-rank because different human motion corresponds to different data latent structure. Figure \ref{Fig.mocap} shows a few examples of the data.

In this paper, we consider the Trial 09 of subject $\#01$ and the Trial 06 of subject $\#56$. The sizes of the corresponding data matrices are $62\times 4242$ and $62\times 6784$, respectively. We add Gaussian noises to $10\%$ or $30\%$ of the entries of the two matrices, where the variance of the noise is the same as that of the data.
In RPCA, the parameter $\lambda$ is set as $1/\sqrt{n}$ or $1.5/\sqrt{n}$; the Lagrange penalty parameter is set as $\lambda$. In RDL, we set $d=31$, $\lambda_C=0.07$ or $0.08$, and $\lambda_E=2$ or $3$. In RNLMF, we set $d=62$, $\sigma=0.5n^{-2}\sum_{ij}\Vert\hat{\bm{x}}_i-\hat{\bm{x}}_j\Vert$, $\lambda_C=0.01$, and choose $\lambda_E$ from $\lbrace 2\times 10^{-5},4\times 10^{-5},5\times 10^{-5}\rbrace$.

Since the variances of the 62 signals are not at the same level, we also consider the normalized mean-absolute-error
$$\textup{MAE}:=\Vert\bm{X}-\hat{\bm{X}}\Vert_1/\Vert\bm{X}\Vert_1.$$

Table \ref{Tab_real_mocap} shows the average RMSE and MAE of 20 repeated trials and the standard deviation. RNLMF  outperformed other methods significantly especially when $\rho=0.3$.

We randomly split the data into two subsets of equal size. We perform RPCA, RDL, and RNLMF on one subset (training data) and then use the trained model to denoise the other subset (testing data). Let $\hat{\bm{X}}'$ be the noisy testing data. For RPCA, we consider the following problem
\begin{equation}
\mathop{\textup{minimize}}_{\bm{V},\bm{E}'}~
\dfrac{1}{2}\Vert\hat{\bm{X}}'-\bm{U}\bm{V}-\bm{E}'\Vert_F^2+\lambda_V\Vert \bm{V}\Vert_F^2+\lambda_E\Vert\bm{E}'\Vert_1,
\end{equation}
where $\bm{U}\in\mathbb{R}^{m\times r}$ consists of the first $r$ left singular vectors of $\bm{X}$ obtained by solving \eqref{Eq.RPCA} and  $r$ is set to be 10 or 20 in this study.
For RDL, we consider
\begin{equation}
\mathop{\textup{minimize}}_{\bm{C}',\bm{E}'}~
\dfrac{1}{2}\Vert\hat{\bm{X}}'-\bm{D}\bm{C}'-\bm{E}'\Vert_F^2+\lambda_C\Vert \bm{C}'\Vert_1+\lambda_E\Vert\bm{E}'\Vert_1,
\end{equation}
where $\bm{D}$ is obtained by solving \eqref{Eq.RCDL_3}. Notice that the out-of-sample extensions of LRR and SSC use the whole data matrix $\bm{X}$ as a dictionary and hence are not efficient, compared to those of RPCA, RDL, and RNLMF. Moreover, the performance of LRR and SSC are similar to those of RPCA and RDL. Therefore, for simplicity, the out-of-sample extensions of LRR and SSC will not be considered in this study.

The results of 20 repeated trials are reported in Table \ref{Tab_real_mocap_ose}. We see that the recovery performance on training data and testing data are similar. In addition, RNLMF is more effective than RPCA and RDL in denoising new data. The results in Table \ref{Tab_real_mocap_ose} are also similar to those of RNLMF in Table \ref{Tab_real_mocap}.  We conclude that the dictionary matrix $\bm{D}$ given by RNLMF can be used to denoise new data efficiently and the denosing accuracy is comparable to that on the training data.

\section{Conclusion}
We have proposed a new method called RNLMF to recover high-rank matrices from sparse noise. We analyzed the underlying meaning of the factorization loss and the regularization terms in the objective function.  RNLMF can be used in robust dicionary learning, denoising, and clustering and is also scalable to large-scale data. Comparative studies on synthetic data and real data verified the superiority of RNLMF.
One interesting finding is that in RNLMF, $\mathcal{R}=\Vert\bm{C}\Vert_F^2$ yields higher recover accuracy, compared to $\mathcal{R}=\Vert\bm{C}\Vert_\ast$ and $\Vert\bm{C}\Vert_1$. The reason has been analyzed in Section \ref{sec.syn}. It is possible that a sparse $\bm{C}$ given by RNLMF is more useful in sparse coding based classification.

%
%

\appendices
\section*{Appendix}
\subsection{Proof for Lemma \ref{lem_kk}}
\begin{proof}
Let $\bar{\bm{x}}=[\bm{x};\sqrt{c}]$.
\begin{equation*}
\begin{aligned}
&\mathcal{K}_\sigma(\bm{x}_i,\bm{x}_j)=\mathcal{K}_\sigma(\bar{\bm{x}}_i,\bar{\bm{x}}_j)\\
&=\exp\left(-\dfrac{1}{2\sigma^2}\left(\Vert \bar{\bm{x}}_i\Vert^2+\Vert \bar{\bm{x}}_j\Vert^2-2\langle \bar{\bm{x}}_i,\bar{\bm{x}}_j\rangle\right)\right)\\
&=\exp\left(-\dfrac{\Vert \bar{\bm{x}}_i\Vert^2+\Vert \bar{\bm{x}}_j\Vert^2}{2\sigma^2}\right)\exp\left(\dfrac{1}{\sigma^2}\langle \bar{\bm{x}}_i,\bar{\bm{x}}_j\rangle\right)\\
&=\exp\left(-\dfrac{\Vert \bar{\bm{x}}_i\Vert^2+\Vert \bar{\bm{x}}_j\Vert^2}{2\sigma^2}\right)\sum_{u=0}^\infty\dfrac{\langle \bar{\bm{x}}_i,\bar{\bm{x}}_j\rangle^u}{\sigma^{2u}u!}\\
&=\exp\left(-\dfrac{\Vert \bar{\bm{x}}_i\Vert^2+\Vert \bar{\bm{x}}_j\Vert^2}{2\sigma^2}\right)\sum_{u=0}^\infty\dfrac{(\bm{x}_i^\top\bm{x}_j+c)^u}{\sigma^{2u}u!}\\
&=\exp\left(-\dfrac{\Vert \bm{x}_i\Vert^2+\Vert\bm{x}_j\Vert^2+2c}{2\sigma^2}\right)\sum_{u=0}^\infty\dfrac{\mathcal{K}_{c,u}(\bm{x}_i,\bm{x}_j)}{\sigma^{2u}u!}.
\end{aligned}
\end{equation*}
\end{proof}

\subsection{Proof for Corollary \ref{cor_Ebound}}
\begin{proof}
Since $d\geq\textup{rank}(\phi_{c,q}(\bm{X}))$, there exist $\bm{D}\in\mathbb{R}^{m\times d}$ and $\bar{\bm{C}}\in\mathbb{R}^{d\times n}$ such that
$$\phi_{c,q}(\bm{X})=\phi_{c,q}(\bm{D})\bar{\bm{C}}.$$
Let $\bm{C}=\bm{S}_D^{-1}\bar{\bm{C}}\bm{S}_X$, then $$\phi_{c,q}(\bm{X})\bm{S}_X=\phi_{c,q}(\bm{D})\bm{S}_D\bm{C}.$$
As $\phi_{c,q}$ contains all features of $\phi_{c,u}$ with $u\leq q$, i.e. $\phi_{c,q}=[\phi_{c,u},\ldots]^\top$, we have
$$\phi_{c,u}(\bm{X})\bm{S}_X=\phi_{c,u}(\bm{D})\bm{S}_D\bm{C},\quad \forall\ 0\leq u\leq q.$$
Combing these equalities with Lemma \ref{lem_objk1k2}, we finish the proof.
\end{proof}

\subsection{Proof for Lemma \ref{lem_objk1k2}}
\begin{proof}
Define $w_u=1/(\sigma^u\sqrt{u!})$. We have
\begin{equation}
\begin{aligned}
&\dfrac{1}{2}\Vert\phi_\sigma(\bm{X})-\phi_\sigma(\bm{D})\bm{C}\Vert_F^2\\
=&\sum_{u=0}^{\infty}\dfrac{w_u^2}{2}\Vert\phi_{c,u}(\bm{X})\bm{S}_X-\phi_{c,u}(\bm{D})\bm{S}_D\bm{C}\Vert_F^2\\
=&\sum_{u=0}^{q}\dfrac{w_u^2}{2}\Vert\phi_{c,u}(\bm{X})\bm{S}_X-\phi_{c,u}(\bm{D})\bm{S}_D\bm{C}\Vert_F^2\\
&+R_1+R_2+R_3,
\end{aligned}
\end{equation}
where $R_1=\sum_{u=q+1}^{\infty}\dfrac{w_u^2}{2}\textup{Tr}\big(\mathcal{K}_{c,u}(\bm{X},\bm{X})\bm{S}_X\bm{S}_X\big)$, $R_2=-\sum_{u=q+1}^{\infty}w_u^2\textup{Tr}\big(\bm{C}^{\top}\bm{S}_D^{\top}\mathcal{K}_{c,u}(\bm{D},\bm{X})\bm{S}_X\big)$, and $R_3=\sum_{u=q+1}^{\infty}\dfrac{w_u^2}{2}(\textup{Tr}\big(\bm{C}^{\top}\bm{S}_D^{\top}\mathcal{K}_{c,u}(\bm{D},\bm{D})\bm{S}_D\bm{C}\big)$. Suppose $\sigma^2>\kappa_2+c$. We have
\begin{equation}
\begin{aligned}
\vert R_1\vert&=\sum_{i=1}^ns_i^2\sum_{u=q+1}^{\infty}\dfrac{w_{u}^2}{2}(\Vert\bm{x}_i\Vert^2+c)^{u}\\
&\leq\sum_{i=1}^n\dfrac{s_i^2}{2(q!)}\Big(\dfrac{\Vert\bm{x}_i\Vert^2+c}{\sigma^2}\Big)^q\\
&\leq\dfrac{0.5n\exp(-\tfrac{c}{\sigma^2})}{q!}\Big(\dfrac{\max_i\Vert\bm{x}_i\Vert^2+c}{\sigma^2}\Big)^{q}.
\end{aligned}
\end{equation}
\begin{equation}
\begin{aligned}
\vert R_2\vert&\leq\sum_{u=q+1}^{\infty}w_u^2\Vert\bm{C}\Vert_F\Vert\bm{S}_D\Vert_2\Vert\bm{S}_X\Vert_2\Vert\mathcal{K}_{c,u}(\bm{D},\bm{X})\Vert_F\\
&\leq\sqrt{dn}\exp(-\tfrac{c}{\sigma^2})\Vert\bm{C}\Vert_F\sum_{u=q+1}^{\infty}w_u^2(\max_{ij}\vert\bm{x}_i^\top\bm{d}_j+c\vert)^{u}\\
&\leq\dfrac{\sqrt{dn}\exp(-\tfrac{c}{\sigma^2})\Vert\bm{C}\Vert_F}{q!}\big(\dfrac{\max_{ij}\Vert\bm{x}_i\Vert\Vert\bm{d}_j\Vert+c}{\sigma^2}\big)^{q}.
\end{aligned}
\end{equation}
\begin{equation}
\begin{aligned}
\vert R_3\vert&\leq\sum_{u=q+1}^{\infty}\dfrac{w_u^2}{2}\Vert\bm{C}\Vert_2\Vert\bm{C}\Vert_F\Vert\bm{S}_D\Vert_2^2\Vert\mathcal{K}_{c,u}(\bm{D},\bm{D})\Vert_F\\
&\leq 0.5d\exp(-\tfrac{c}{\sigma^2})\Vert\bm{C}\Vert_2\Vert\bm{C}\Vert_F\sum_{u=q+1}^{\infty}w_u^2(\max_{ij}\vert\bm{d}_i^\top\bm{d}_j+c\vert)^{u}\\
&\leq \dfrac{0.5d\exp(-\tfrac{c}{\sigma^2})\Vert\bm{C}\Vert_2\Vert\bm{C}\Vert_F}{q!}\big(\dfrac{\max_{i}\Vert\bm{d}_i\Vert^2+c}{\sigma^2}\big)^{q}.
\end{aligned}
\end{equation}
Let $\kappa_1=\max\lbrace 0.5n,\sqrt{dn}\Vert\bm{C}\Vert_F,0.5d\Vert\bm{C}\Vert_2\Vert\bm{C}\Vert_F\rbrace$ and $\kappa_2=\max\lbrace\max_{i}\Vert\bm{x}_i\Vert^2,\max_{j}\Vert\bm{d}_j\Vert^2\rbrace$. We obtain
\begin{equation}
\begin{aligned}
\vert R_1\vert+\vert R_2\vert+\vert R_3\vert\leq \dfrac{3\kappa_1\exp(-\tfrac{c}{\sigma^2})}{q!}\big(\dfrac{\kappa_2+c}{\sigma^2}\big)^{q}.
\end{aligned}
\end{equation}

\end{proof}

\subsection{Proof for Lemma \ref{lem_lbound_0}}

\begin{proof}
It is known that
\begin{equation}\label{eq2_proof_th1}
\min_{\bm{B}\bm{C}=\phi(\bm{X})}\dfrac{1}{2}\Vert\bm{B}\Vert_F^2+\dfrac{1}{2}\Vert\bm{C}\Vert_F^2=\Vert \phi(\bm{X})\Vert_\ast.
\end{equation}
Considering one more constraint $\bm{B}=\phi(\bm{D})$, we must have
\begin{equation}\label{eq2_proof_th2}
\begin{aligned}
&\min_{\bm{B}\bm{C}=\phi(\bm{X}),\bm{B}=\phi(\bm{D})}\dfrac{1}{2}\Vert\phi(\bm{D})\Vert_F^2+\dfrac{1}{2}\Vert\bm{C}\Vert_F^2\\
\geq&\min_{\bm{B}\bm{C}=\phi(\bm{X})}\dfrac{1}{2}\Vert\bm{B}\Vert_F^2+\dfrac{1}{2}\Vert\bm{C}\Vert_F^2.
\end{aligned}
\end{equation}
Combining \eqref{eq2_proof_th1} and \eqref{eq2_proof_th2} finishes the proof.
\end{proof}

\subsection{Proof for Lemma \ref{cor_rbfFbound}}

\begin{proof}
For all $c>0$, we have
$$\Vert\phi(\bm{D})\Vert_F^2+c\Vert\bm{C}\Vert_F^2\geq 2\sqrt{c}\Vert\phi(\bm{D})\bm{C}\Vert_\ast.$$
Choosing $c=\Vert\phi(\bm{D})\Vert_F^2/\Vert\bm{C}\Vert_F^2$, we have
$$\Vert\bm{C}\Vert_F\Vert\phi(\bm{D})\Vert_F\geq \Vert\phi(\bm{D})\bm{C}\Vert_\ast.$$
Recalling $\Vert\phi(\bm{D})\Vert_F^2\equiv d$, we arrive at
$$\Vert\bm{C}\Vert_F\geq \Vert\phi(\bm{D})\bm{C}\Vert_\ast/\sqrt{d}.$$
\end{proof}

\subsection{Proof for Lemma \ref{lem_LC}}

\begin{proof}
Note that
$\nabla_{\bm{C}}\mathcal{L}(\bm{C})=-\mathcal{K}(\bm{D}_{t-1},\hat{\bm{X}}-\bm{E}_{t-1})+\mathcal{K}(\bm{D}_{t-1},\bm{D}_{t-1})\bm{C}$ is $L_C^t$-Lipschitz continuous, where $L_C^t=\Vert\mathcal{K}(\bm{D}_{t-1},\bm{D}_{t-1})\Vert_2$. Thus
\begin{equation}\label{Eq.proof_CLip}
\begin{aligned}
\mathcal{L}(\bm{C}_t)\leq\mathcal{L}(\bm{C}_{t-1})&+\langle \bm{C}_t-\bm{C}_{t-1}, \nabla_{\bm{C}}\mathcal{L}(\bm{C}_{t-1})\rangle\\
&+\dfrac{L_C^t}{2}\Vert \bm{C}_t-\bm{C}_{t-1}\Vert_F^2.
\end{aligned}
\end{equation}
According to the definition of the proximal map $\Theta_u$ (or $\Psi_u$) \cite{parikh2014proximal}, we have
\begin{equation}\label{Eq.proof_CProx}
\begin{aligned}
\bm{C}_t\in \min\limits_{\bm{C}}\ &\langle \bm{C}-\bm{C}_{t-1}, \nabla_{\bm{C}}\mathcal{L}(\bm{C}_{t-1})\rangle\\
&+\dfrac{\tau_C^t}{2}\Vert \bm{C}-\bm{C}_{t-1}\Vert_F^2+\lambda_C\mathcal{R}(\bm{C}).
\end{aligned}
\end{equation}
where $\mathcal{R}(\bm{C})=\Vert\bm{C}\Vert_1$ (or $\Vert\bm{C}\Vert_\ast$).
By taking $\bm{C}=\bm{C}_{t-1}$, it follows from \eqref{Eq.proof_CProx} that
\begin{equation}\label{Eq.proof_CProx1}
\begin{aligned}
&\langle \bm{C}_t-\bm{C}_{t-1}, \nabla_{\bm{C}}\mathcal{L}(\bm{C}_{t-1})\rangle+\lambda_C\mathcal{R}(\bm{C}_t)\\
\leq&\lambda_C\mathcal{R}(\bm{C}_{t-1})-\dfrac{\tau_C^t}{2}\Vert \bm{C}_t-\bm{C}_{t-1}\Vert^2_F.
\end{aligned}
\end{equation}
Combining \eqref{Eq.proof_CLip} and \eqref{Eq.proof_CProx1}, we have
\begin{equation*}
\begin{aligned}
\mathcal{L}(\bm{C}_{t})+\lambda_C\mathcal{R}(\bm{C}_{t})\leq &\mathcal{L}(\bm{C}_{t-1})+\lambda_C\mathcal{R}(\bm{C}_{t-1})\\
&-\dfrac{\tau_C^t-L_C^t}{2}\Vert \bm{C}_t-\bm{C}_{t-1}\Vert^2_F.
\end{aligned}
\end{equation*}
This finished the proof.
\end{proof}

\subsection{Proof for Lemma \ref{lem_LD}}

\begin{proof}
Recall that $
\mathcal{L}(\bm{D})=-\textup{Tr}\left(\bm{C}_t^{\top}\mathcal{K}(\bm{D},\hat{\bm{X}}-\bm{E}_{t-1})\right)+\dfrac{1}{2}\textup{Tr}\Big(\bm{C}_t^{\top}\mathcal{K}(\bm{D},\bm{D})\bm{C}_t\Big)$
and the gradient is
\begin{equation*}
\begin{aligned}
\nabla_{\bm{D}}\mathcal{L}(\bm{D})=&\tfrac{1}{\sigma^2}(\hat{\bm{X}}-\bm{E}_{t-1})\bm{W}_1^t-\tfrac{1}{\sigma^2}\bm{D}\bar{\bm{W}}_1^t\\
&+\tfrac{2}{\sigma^2}\bm{D}\bm{W}_2^t-\tfrac{2}{\sigma^2}\bm{D}\bar{\bm{W}}_2^t,
\end{aligned}
\end{equation*}
where $\bm{W}_1^t=-\bm{C}_{t}^\top\odot\mathcal{K}(\hat{\bm{X}}-\bm{E}_{t-1},\bm{D})$, $\bm{W}_2^t=(0.5\bm{C}_t\bm{C}_t^\top)\odot\mathcal{K}(\bm{D},\bm{D})$, $\bar{\bm{W}}_1^t=\mbox{diag}(\bm{1}_n^\top\bm{W}_1^t)$, and $\bar{\bm{W}}_2=\mbox{diag}(\bm{1}_d^\top\bm{W}_2)$.
Let $s_{ij}=\exp(-\tfrac{\Vert [\hat{\bm{X}}-\bm{E}]_{:i}\Vert^2+\Vert[\bm{D}]_{:j}\Vert^2}{2\sigma^2})$ and suppose $\sigma$ is large enough. Then according to Lemma \ref{lem_kk}, we have
$$ \bm{W}_1^t\approx-\bm{C}_t^\top\odot\bm{S}\odot(1+\tfrac{(\hat{\bm{X}}-\bm{E})^\top\bm{D}}{\sigma^2}),$$
where we have omitted the higher ($u>2$) order terms of the polynomial approximate of Gaussian RBF kernel. Similarly, we have
$$ \bm{W}_2^t\approx(0.5\bm{C}_t\bm{C}_t^\top)\odot\bm{S}_D\odot(1+\tfrac{\bm{D}^\top\bm{D}}{\sigma^2}),$$
where $[\bm{S}_D]_{ij}=\exp(-\tfrac{\Vert [\bm{D}]_{:i}\Vert^2+\Vert[\bm{D}]_{:j}\Vert^2}{2\sigma^2})$.

Let's check the sensitivity of $\nabla_{\bm{D}}\mathcal{L}(\bm{D})$ to the perturbation on $\bm{D}$. First, consider the first term in $\nabla_{\bm{D}}\mathcal{L}(\bm{D})$ and let $[\hat{\bm{S}}]_{ij}=\exp(-\tfrac{\Vert [\hat{\bm{X}}-\bm{E}]_{:i}\Vert^2+\Vert[\hat{\bm{D}}]_{:j}\Vert^2}{2\sigma^2})$. We have
\begin{align}\label{eq.DDD1}
&\Vert\tfrac{1}{\sigma^2}(\hat{\bm{X}}-\bm{E}_{t-1})(\bm{C}_t^\top\odot(\bm{S}\odot(1+\tfrac{(\hat{\bm{X}}-\bm{E})^\top\bm{D}}{\sigma^2}) \nonumber \\ \nonumber
&\quad-\hat{\bm{S}}\odot(1+\tfrac{(\hat{\bm{X}}-\bm{E})^\top\hat{\bm{D}}}{\sigma^2})))\Vert_F \nonumber \\ \nonumber
\leq&\tfrac{1}{\sigma^2}\Vert\hat{\bm{X}}-\bm{E}_{t-1}\Vert \Vert\bm{C}_t\Vert_\infty\max\lbrace 2\Vert\bm{S}\Vert_\infty,2\Vert\hat{\bm{S}}\Vert_\infty\rbrace \nonumber\\ \nonumber
&\quad \times \Vert \sigma^{-2} (\hat{\bm{X}}-\bm{E}_{t-1})(\bm{D}-\hat{\bm{D}})\Vert_F\\
\leq &\tfrac{ 2\Vert\bm{C}_t\Vert_\infty}{\sigma^4}\Vert\hat{\bm{X}}-\bm{E}_{t-1}\Vert^2\Vert\bm{D}-\bm{D}\Vert_F,
\end{align}
where $\Vert\bm{S}\Vert_\infty$,$\Vert\hat{\bm{S}}\Vert_\infty\leq 1$. We see that when $\sigma$ is large and $\Vert\bm{C}\Vert_\infty$ is small, the first term in $\nabla_{\bm{D}}\mathcal{L}(\bm{D})$ is not sensitive to the changes of $\bm{D}$.

Now consider the third term of $\nabla_{\bm{D}}\mathcal{L}(\bm{D})$. Let $\hat{\bm{W}}_2^t$ be the perturbed copy of $\bm{W}_2^t$ computed from $\hat{\bm{D}}$. We have
\begin{align}\label{eq.DDD}
&\tfrac{2}{\sigma^2}\Vert\bm{D}\bm{W}_2^t-\hat{\bm{D}}\hat{\bm{W}}_2^t\Vert_F \nonumber \\ \nonumber
\approx&\tfrac{2}{\sigma^2}\Vert(\bm{D}-\hat{\bm{D}})\bm{W}_2^t-\hat{\bm{D}}((0.5\bm{C}_t\bm{C}_t^\top)\odot\bm{S}_D\odot(1+\tfrac{\bm{D}^\top\bm{D}}{\sigma^2})\\ \nonumber
&\quad-(0.5\bm{C}_t\bm{C}_t^\top)\odot\hat{\bm{S}}_D\odot(1+\tfrac{\hat{\bm{D}}^\top\hat{\bm{D}}}{\sigma^2}))\Vert_F \nonumber\\
\leq & \tfrac{2}{\sigma^2}\Vert\bm{W}_2^t\Vert\Vert\bm{D}-\hat{\bm{D}}\Vert_F+\tfrac{2}{\sigma^4}\Vert \hat{ \bm{D}} \Vert\Vert \bm{C}_t\bm{C}_t^\top\Vert_\infty\Vert \bm{D}^\top\bm{D}-\hat{\bm{D}}^\top\hat{\bm{D}}\Vert_F  \nonumber \\ \nonumber
\leq & \tfrac{2}{\sigma^2}\Vert\bm{W}_2^t\Vert\Vert\bm{D}-\hat{\bm{D}}\Vert_F\\
&\quad+\tfrac{2}{\sigma^4}(\Vert \hat{\bm{D}} \Vert^2+\Vert \bm{D} \Vert\Vert \hat{\bm{D}} \Vert)\Vert \bm{C}_t\bm{C}_t^\top\Vert_\infty\Vert \bm{D}-\hat{\bm{D}}\Vert_F.
\end{align}
In \eqref{eq.DDD}, when $\sigma$ is large enough,  the second term can be smaller than the first term and $\bm{W}_2^t\approx0.5\bm{C}_t\bm{C}_t^\top$. It means the $\bm{D}$ in $\bm{W}_2^t$ makes small contribution to $\bm{D}\bm{W}_2^t$.  Therefore, the contribution of $\bm{D}$ in $\bm{W}_2^t$  to the Hessian of $\mathcal{L}(\bm{D})$ can be neglected. The conclusion also applies to  $\bar{\bm{W}}_2^t$ and $\bar{\bm{W}}_1^t$. In addition,  comparing \eqref{eq.DDD1} with \eqref{eq.DDD}, we can neglect the contribution of the first term of $\nabla_{\bm{D}}\mathcal{L}(\bm{D})$ to the Hessian of $\mathcal{L}(\bm{D})$.

Now let's consider second order approximation of $\mathcal{L}(D)$ around $\bm{D}_{t-1}$:
\begin{equation}\label{Eq.LDeqH}
\begin{aligned}
\mathcal{L}(\bm{D})\approx&\mathcal{L}(\bm{D}_{t-1})+\langle\nabla_{\bm{D}}\mathcal{L}(\bm{D}_{t-1}), \bm{D}-\bm{D}_{t-1}\rangle\\
&+\tfrac{1}{2}\text{vec}(\bm{D}-\bm{{D}_{t-1}})^\top\bm{\mathcal{H}}_{t-1}\textup{vec}(\bm{D}-\bm{{D}_{t-1}})^\top,
\end{aligned}
\end{equation}
where $\bm{\mathcal{H}}_{t-1}$ is the Hessian at iteration $t-1$. It is difficult to compute $\bm{\mathcal{H}}_{t-1}$. However, out previous analysis indicates that we can treat $\bm{W}_1^t$, $\bar{\bm{W}}_1^t$, $\bm{W}_2^t$, and $\bar{\bm{W}}_2^t$ as constants independent of $\bm{D}$ at iteration $t$. Thus the estimated Hessian is
$$\hat{\bm{\mathcal{H}}}_{t-1}=\left[\begin{matrix}
\bm{H}_{t-1} &\ldots& \bm{0}\\
\vdots &\ddots & \vdots\\
\bm{0} &\ldots & \bm{H}_{t-1}
\end{matrix}
\right]\in\mathbb{R}^{md\times md}
$$
where $\bm{H}_{t-1}=\tfrac{1}{\sigma^2}(-\bar{\bm{W}}_1^t+2\bm{W}_2^t-2\bar{\bm{W}}_2^t)\in\mathbb{R}^{d\times d}$.

Let $\mu\geq 0$ be  sufficiently large such that $\bm{H}_{t-1}+\mu\bm{I}$ is positive definite. Let $\tau_D^t\geq1$ be sufficiently large such that
\begin{equation}\label{Eq.LDineq}
\begin{aligned}
\mathcal{L}(\bm{D})\leq&\mathcal{L}(\bm{D}_{t-1})+\langle\nabla_{\bm{D}}\mathcal{L}(\bm{D}_{t-1}), \bm{D}-\bm{D}_{t-1}\rangle\\
&+\tfrac{\tau_D^t}{2}\text{Tr}\left((\bm{D}-\bm{{D}_{t-1}})(\bm{H}_{t-1}+\mu\bm{I})(\bm{D}-\bm{{D}_{t-1}})^\top\right).
\end{aligned}
\end{equation}
We then minimize the right side of \eqref{Eq.LDineq}  by letting the derivative be zero and get
\begin{equation}\label{Eq.LDineq_sol}
\bm{D}_t=\bm{D}_{t-1}-\dfrac{1}{\tau_D^t}\nabla_{\bm{D}}\mathcal{L}(\bm{D}_{t-1})(\bm{H}_{t-1}+\mu\bm{I})^{-1}.
\end{equation}
Invoking \eqref{Eq.LDineq_sol} into \eqref{Eq.LDineq} yields
\begin{equation*}
\begin{aligned}
&\mathcal{L}(\bm{D}_t)\leq \mathcal{L}(\bm{D}_{t-1})\\
&-\dfrac{1}{2\tau_D^t}\text{Tr}\left(\nabla_{\bm{D}}\mathcal{L}(\bm{D}_{t-1})(\bm{H}_{t-1}+\mu\bm{I})^{-1}\nabla_{\bm{D}}\mathcal{L}(\bm{D}_{t-1})^\top\right).
\end{aligned}
\end{equation*}
Since $\bm{H}_{t-1}+\mu\bm{I}$ is positive definite, we have $$\mathcal{L}(\bm{D}_t)-\mathcal{L}(\bm{D}_{t-1})\leq 0.$$
This finished the proof.
\end{proof}

\subsection{Proof for Corollary \ref{cor_LD}}
\begin{proof}
Recall that
$\bm{D}_{t}=\bm{D}_{t-1}-\bm{\Delta}_t$,
where $\bm{\Delta}_t=\eta\bm{\Delta}_{t-1}+\tfrac{1}{\tau_D^t}\nabla_{\bm{D}}\mathcal{L}(\bm{D}_{t-1})(\bm{H}_{t-1}+\mu\bm{I})^{-1}$ and $0<\eta<1$.
Follow the same analysis on $\bm{H}$ in the proof of Lemma \ref{lem_LD}.
In \eqref{Eq.LDineq}, letting $\bm{D}=\bm{D}_t$, we have
\begin{equation*}
\begin{aligned}
&\mathcal{L}(\bm{D}_t)\\
\leq& \mathcal{L}(\bm{D}_{t-1})-\langle\nabla_{\bm{D}}\mathcal{L}(\bm{D}_{t-1}), \eta\bm{\Delta}_{t-1}\rangle\\
&-\langle\nabla_{\bm{D}}\mathcal{L}(\bm{D}_{t-1}), \tfrac{1}{\tau_D^t}\nabla_{\bm{D}}\mathcal{L}(\bm{D}_{t-1})(\bm{H}_{t-1}+\mu\bm{I})^{-1}\rangle\\
&+\tfrac{\tau_D^t}{2}\text{Tr}\left(\eta^2\bm{\Delta}_{t-1}(\bm{H}_{t-1}+\mu\bm{I})bm{\Delta}_{t-1}^\top\right)\\
&+\tfrac{\tau_D^t}{2}\text{Tr}\left(\eta\bm{\Delta}_{t-1}(\bm{H}_{t-1}+\mu\bm{I})(\tfrac{1}{\tau_D^t}\nabla_{\bm{D}}\mathcal{L}(\bm{D}_{t-1})(\bm{H}_{t-1}+\mu\bm{I})^{-1})^\top\right)\\
&+\tfrac{\tau_D^t}{2}\text{Tr}\left((\tfrac{1}{\tau_D^t}\nabla_{\bm{D}}\mathcal{L}(\bm{D}_{t-1})(\bm{H}_{t-1}+\mu\bm{I})^{-1})(\bm{H}_{t-1}+\mu\bm{I})\eta\bm{\Delta}_{t-1}^\top\right)\\
&+\tfrac{1}{2\tau_D^t}\text{Tr}\left(\nabla_{\bm{D}}\mathcal{L}(\bm{D}_{t-1})(\bm{H}_{t-1}+\mu\bm{I})^{-1}\nabla_{\bm{D}}\mathcal{L}(\bm{D}_{t-1})^\top\right)\\
=&\mathcal{L}(\bm{D}_{t-1})-\tfrac{1}{2\tau_D^t}\text{Tr}\left(\nabla_{\bm{D}}\mathcal{L}(\bm{D}_{t-1})(\bm{H}_{t-1}+\mu\bm{I})^{-1}\nabla_{\bm{D}}\mathcal{L}(\bm{D}_{t-1})^\top\right)\\
&-\langle\nabla_{\bm{D}}\mathcal{L}(\bm{D}_{t-1}), \eta\bm{\Delta}_{t-1}\rangle\\
&+\tfrac{\eta^2\tau_D^t}{2}\text{Tr}\left(\bm{\Delta}_{t-1}(\bm{H}_{t-1}+\mu\bm{I})\bm{\Delta}_{t-1}^\top\right)\\
&+\eta\text{Tr}\left(\bm{\Delta}_{t-1}\nabla_{\bm{D}}\mathcal{L}(\bm{D}_{t-1})^\top\right)\\
=&\mathcal{L}(\bm{D}_{t-1})-\tfrac{1}{2\tau_D^t}\text{Tr}\left(\nabla_{\bm{D}}\mathcal{L}(\bm{D}_{t-1})(\bm{H}_{t-1}+\mu\bm{I})^{-1}\nabla_{\bm{D}}\mathcal{L}(\bm{D}_{t-1})^\top\right)\\
&+\tfrac{\eta^2\tau_D^t}{2}\text{Tr}\left(\bm{\Delta}_{t-1}(\bm{H}_{t-1}+\mu\bm{I})\bm{\Delta}_{t-1}^\top\right)\\
\end{aligned}
\end{equation*}

\end{proof}

\subsection{Proof for Lemma \ref{lem_LE}}

\begin{proof}
Note that
\begin{equation*}
\nabla_{\bm{E}}\mathcal{L}(\bm{E})=\tfrac{1}{\sigma^2}\big((\hat{\bm{X}}-\bm{E})\bar{\bm{W}}_3^t-\bm{D}_t\bm{W}_3^t\big),
\end{equation*}
where $\bm{W}_3^t=-\bm{C}_{t}\odot\mathcal{K}(\bm{D}_t,\hat{\bm{X}}-\bm{E})$ and $\bar{\bm{W}}_3^t=\mbox{diag}(\bm{1}_d^\top\bm{W}_3^t)$.
According to Lemma \ref{lem_kk} (let $c=0$ and assume $\sigma$ is large enough), we have
$$ [\bm{W}_3^t]_{ij}\approx-[\bm{C}_t]_{ij}s_{ij}(1+\tfrac{[\bm{D}_t]_{:i}^\top(\bm{x}_j-\bm{e}_j)}{\sigma^2})\approx -[\bm{C}_t]_{ij},$$
where $s_{ij}=\exp(-\tfrac{\Vert[\bm{D}_t]_{:i}\Vert^2+\Vert \bm{x}_j-\bm{e}_j\Vert^2}{2\sigma^2})$. Let $\hat{\bm{E}}$ be an perturbed copy of $\bm{E}$. It follows that
\begin{align*}
&\Vert\bm{W}_3^t-\hat{\bm{W}}_3^t\Vert_F\\
\approx&\Vert\bm{C}_t\odot\bm{S}\odot(1+\tfrac{\bm{D}_t^\top(\bm{X}-\bm{E})}{\sigma^2})-\bm{C}_t\odot\hat{\bm{S}}\odot(1+\tfrac{\bm{D}_t^\top(\bm{X}-\hat{\bm{E}})}{\sigma^2})\Vert_F\\
\leq &2\sigma^{-2}\Vert\bm{C}\Vert_\infty\max\lbrace\Vert\bm{S}\Vert_\infty,\Vert\hat{\bm{S}}\Vert_\infty\rbrace\Vert\bm{D}_t^\top(\bm{E}-\hat{\bm{E}})\Vert_F\\
\leq &2\sigma^{-2}\Vert\bm{C}\Vert_\infty\Vert\bm{D}_t\Vert\Vert\bm{E}-\hat{\bm{E}}\Vert_F.
\end{align*}
In addition, $\Vert \bar{\bm{W}}_3^t-\hat{\bar{\bm{W}}}_3^t\Vert_F\leq\sqrt{d}\Vert\bm{W}_3^t-\hat{\bm{W}}_3^t\Vert_F$.

Then we have
\begin{equation}
\begin{aligned}
&\Vert\nabla_{\bm{E}}\mathcal{L}(\bm{E})-\nabla_{\hat{\bm{E}}}\mathcal{L}(\hat{\bm{E}})\Vert_F\\
\leq &\tfrac{1}{\sigma^2}\Vert(\hat{\bm{X}}-\bm{E})\bar{\bm{W}}_3^t-\bm{D}_t\bm{W}_3^t-(\hat{\bm{X}}-\bm{E})\hat{\bar{\bm{W}}}_3^t-\bm{D}_t\hat{\bm{W}}_3^t\Vert_F\\
&+\Vert(\hat{\bm{X}}-\bm{E})\hat{\bar{\bm{W}}}_3^t-\bm{D}_t\hat{\bm{W}}_3^t-(\hat{\bm{X}}-\hat{\bm{E}})\hat{\bar{\bm{W}}}_3^t-\bm{D}_t\hat{\bm{W}}_3^t\Vert_F\\
\leq & \tfrac{1}{\sigma^2}\Vert\hat{\bm{X}}-\bm{E}\Vert\Vert \bar{\bm{W}}_3^t-\hat{\bar{\bm{W}}}_3^t\Vert_F+\tfrac{1}{\sigma^2}\Vert
\bm{D}_t\Vert\Vert\bm{W}_3^t-\hat{\bm{W}}_3^t\Vert_F\\
&+\tfrac{1}{\sigma^2}\Vert\hat{\bar{\bm{W}}}_3^t\Vert\Vert\bm{E}-\hat{\bm{E}}\Vert_F\\
\lessapprox &(\tfrac{2\Vert \bm{C}\Vert_\infty}{\sigma^4}(\sqrt{d}\Vert\hat{\bm{X}}-\bm{E}\Vert\Vert\bm{D}_t\Vert+\Vert
\bm{D}_t\Vert^2)+\tfrac{1}{\sigma^2}\Vert\hat{\bar{\bm{W}}}_3^t\Vert)\Vert\bm{E}-\hat{\bm{E}}\Vert_F\\
\leq &\tfrac{\xi }{\sigma^2}\Vert\hat{\bar{\bm{W}}}_3^t\Vert\Vert\bm{E}-\hat{\bm{E}}\Vert_F,
\end{aligned}
\end{equation}
where $\xi\geq 1$ is a large enough constant. Since $\sigma$ is sufficiently large and $\Vert\bm{C}\Vert_\infty$ is encouraged to be small in its update, $\xi$ will not be too large.

We may estimate the Lipschitz constant of $\nabla_{\bm{E}}\mathcal{L}(\bm{E})$ as
$$\hat{L}_E^t=\tfrac{2\Vert \bm{C}\Vert_\infty}{\sigma^4}(\sqrt{d}\Vert\hat{\bm{X}}-\bm{E}\Vert\Vert\bm{D}_t\Vert+\Vert
\bm{D}_t\Vert^2)+\tfrac{1}{\sigma^2}\Vert\hat{\bar{\bm{W}}}_3^t\Vert,$$
which however will increase the computational cost because of the spectral norms of $\hat{\bm{X}}-\bm{E}$ and $\bm{D}_t$. For simplicity, we use
$$\hat{L}_E^t=\xi\Vert\bar{\bm{W}}_3^t\Vert/\sigma^2=\xi\Vert\bm{1}_d^\top\bm{W}_3^t\Vert_\infty/\sigma^2,$$
where $\xi\geq 1$ should be large enough.
Then letting $\tau_E^t> \hat{L}_E^t$, we have
\begin{equation}\label{Eq.proof_ELip}
\begin{aligned}
\mathcal{L}(\bm{E})\leq\mathcal{L}(\bm{E}_{t-1})&+\langle \bm{E}-\bm{E}_{t-1}, \nabla_{\bm{E}}\mathcal{L}(\bm{E}_{t-1})\rangle\\
&+\dfrac{\tau_E^t}{2}\Vert \bm{E}-\bm{E}_{t-1}\Vert_F^2,
\end{aligned}
\end{equation}
provided that $\xi$ is sufficiently large.
According to Table \ref{Tab_sol_E} and the definition of the proximal map \cite{parikh2014proximal}, we have
\begin{equation}\label{Eq.proof_EProx}
\begin{aligned}
\bm{E}_t\in \min\limits_{\bm{E}}\ &\langle \bm{E}-\bm{E}_{t-1}, \nabla_{\bm{E}}\mathcal{L}(\bm{E}_{t-1})\rangle\\
&+\dfrac{\tau_E^t}{2}\Vert \bm{E}-\bm{E}_{t-1}\Vert_F^2+\lambda_E\mathcal{R}(\bm{E}).
\end{aligned}
\end{equation}
where $\mathcal{R}(\bm{E})=\Vert\bm{E}\Vert_F^2$, $\Vert\bm{E}\Vert_1$, or $\Vert\bm{E}\Vert_{2,1}$.
By taking $\bm{E}=\bm{E}_{t-1}$, it follows from \eqref{Eq.proof_EProx} that
\begin{equation}\label{Eq.proof_EProx1}
\begin{aligned}
&\langle \bm{E}_t-\bm{E}_{t-1}, \nabla_{\bm{E}}\mathcal{L}(\bm{E}_{t-1})\rangle+\lambda_E\mathcal{R}(\bm{E}_t)\\
\leq&\lambda_E\mathcal{R}(\bm{E}_{t-1})-\dfrac{\tau_E^t}{2}\Vert \bm{E}_t-\bm{E}_{t-1}\Vert^2_F.
\end{aligned}
\end{equation}
In addition, the Lipschitz continuity  of $\nabla_{\bm{E}}\mathcal{L}(\bm{E})$ around $\bm{E}_{t-1}$ indicates that
\begin{equation}\label{Eq.proof_ELip_0}
\begin{aligned}
\mathcal{L}(\bm{E}_t)\leq\mathcal{L}(\bm{E}_{t-1})&+\langle \bm{E}_t-\bm{E}_{t-1}, \nabla_{\bm{E}}\mathcal{L}(\bm{E}_{t-1})\rangle\\
&+\dfrac{L_E^t}{2}\Vert \bm{E}_t-\bm{E}_{t-1}\Vert_F^2,
\end{aligned}
\end{equation}
Combining \eqref{Eq.proof_ELip_0} and \eqref{Eq.proof_EProx1}, we have
\begin{equation*}
\begin{aligned}
\mathcal{L}(\bm{E}_{t})+\lambda_E\mathcal{R}(\bm{E}_{t})\leq &\mathcal{L}(\bm{E}_{t-1})+\lambda_E\mathcal{R}(\bm{E}_{t-1})\\
&-\dfrac{\tau_E^t-L_E^t}{2}\Vert \bm{E}_t-\bm{E}_{t-1}\Vert^2_F.
\end{aligned}
\end{equation*}
Let $\tau_E^t=\xi\Vert\bm{1}_d^\top\bm{W}_3^t\Vert_\infty/\sigma^2$ and $\xi$ be sufficiently large, then $\tau_E^t-L_E^t\geq 0$.
This finished the proof.
\end{proof}

\subsection{Proof for Theorem \ref{Eq.converge}}
\begin{proof}
Combining Lemmas \ref{lem_LC}, \ref{lem_LD}, and \ref{lem_LE}, we have
\begin{equation}\label{Eq.delta_J}
\mathcal{J}(\bm{D}_t,\bm{C}_t,\bm{E}_t)\leq \mathcal{J}(\bm{D}_{t-1},\bm{C}_{t-1},\bm{E}_{t-1})-\Delta_J^t,
\end{equation}
where $\Delta_J^t=\Delta_{J_C}^t+\Delta_{J_D}^t+\Delta_{J_E}^t$ and
\begin{align}
\Delta_{J_C}^t=&\dfrac{\tau_C^t-L_C^t}{2}\Vert \bm{C}_t-\bm{C}_{t-1}\Vert_F^2 \nonumber,\\
\Delta_{J_D}^t=&\dfrac{1}{2\tau_D^t}\textup{Tr}\left(\nabla_{\bm{D}}\mathcal{L}(\bm{D}_{t-1})(\bm{H}_{t-1}+\mu\bm{I})^{-1}\nabla_{\bm{D}}\mathcal{L}(\bm{D}_{t-1})^\top\right) \nonumber,\\
\Delta_{J_E}^t=&\dfrac{\tau_E^t-L_E^t}{2}\Vert \bm{E}_t-\bm{E}_{t-1}\Vert_F^2. \nonumber
\end{align}
As $\tau_C^t>L_C^t$, $\bm{H}_{t-1}+\mu\bm{I}$ is positive definite, and $\tau_E^t>L_E^t$, we have $\Delta_{J_C}^t\geq 0$,  $\Delta_{J_D}^t\geq 0$, and $\Delta_{J_E}^t\geq 0$. It follows that
$$\mathcal{J}(\bm{D}_t,\bm{C}_t,\bm{E}_t)\leq \mathcal{J}(\bm{D}_{t-1},\bm{C}_{t-1},\bm{E}_{t-1}).$$
Since  $\mathcal{J}(\bm{D},\bm{C},\bm{E})$ is bounded below, we have
$$\lim_{t\rightarrow\infty} \mathcal{J}(\bm{D}_t,\bm{C}_t,\bm{E}_t)-\mathcal{J}(\bm{D}_{t-1},\bm{C}_{t-1},\bm{E}_{t-1})=0.$$

Summing \eqref{Eq.delta_J} from $0$ to $\infty$, we have
\begin{equation*}
\mathcal{J}(\bm{D}_0,\bm{C}_0,\bm{E}_0)-\mathcal{J}(\bm{D}_\infty,\bm{C}_\infty,\bm{E}_\infty)\geq \sum_{t=0}^{\infty} \Delta_J^t.
\end{equation*}
Hence
$$\sum_{t=0}^{\infty} \Delta_{J_C}^t+\Delta_{J_D}^t+\Delta_{J_E}^t<\infty.$$
As $\Delta_{J_C}^t, \Delta_{J_D}^t, \Delta_{J_E}^t\geq 0$ for all $t$, we conclude that
\begin{equation*}
\begin{aligned}
&\lim_{t\rightarrow\infty}\Vert \nabla_{\bm{D}}\mathcal{L}(\bm{D}_{t-1})\Vert_F=0,\\
&\lim_{t\rightarrow\infty}\Vert \bm{C}_t-\bm{C}_{t-1}\Vert_F=0,\\
&\lim_{t\rightarrow\infty}\Vert \bm{E}_t-\bm{E}_{t-1}\Vert_F=0.
\end{aligned}
\end{equation*}
Combining the first equality above with \eqref{Eq.LDineq_sol} yields
$$\lim_{t\rightarrow\infty}\Vert \bm{D}_t-\bm{D}_{t-1}\Vert_F=0.$$
This finished the proof.
\end{proof}
\bibliographystyle{IEEEtran}
\bibliography{Ref_RNLMF}

\begin{thebibliography}{10}
\providecommand{\url}[1]{#1}
\csname url@samestyle\endcsname
\providecommand{\newblock}{\relax}
\providecommand{\bibinfo}[2]{#2}
\providecommand{\BIBentrySTDinterwordspacing}{\spaceskip=0pt\relax}
\providecommand{\BIBentryALTinterwordstretchfactor}{4}
\providecommand{\BIBentryALTinterwordspacing}{\spaceskip=\fontdimen2\font plus
\BIBentryALTinterwordstretchfactor\fontdimen3\font minus
  \fontdimen4\font\relax}
\providecommand{\BIBforeignlanguage}[2]{{%
\expandafter\ifx\csname l@#1\endcsname\relax
\typeout{** WARNING: IEEEtran.bst: No hyphenation pattern has been}%
\typeout{** loaded for the language `#1'. Using the pattern for}%
\typeout{** the default language instead.}%
\else
\language=\csname l@#1\endcsname
\fi
#2}}
\providecommand{\BIBdecl}{\relax}
\BIBdecl

\bibitem{udell2019big}
M.~Udell and A.~Townsend, ``Why are big data matrices approximately low rank?''
  \emph{SIAM Journal on Mathematics of Data Science}, vol.~1, no.~1, pp.
  144--160, 2019.

\bibitem{wold1987principal}
S.~Wold, K.~Esbensen, and P.~Geladi, ``Principal component analysis,''
  \emph{Chemometrics and intelligent laboratory systems}, vol.~2, no. 1-3, pp.
  37--52, 1987.

\bibitem{PCA_Jolliffe2002}
I.~Jolliffe, \emph{Principal Component Analysis}, ser. Springer Series in
  Statistics.\hskip 1em plus 0.5em minus 0.4em\relax Springer, 2002.

\bibitem{RPCA}
E.~J. Cand\`{e}s, X.~Li, Y.~Ma, and J.~Wright, ``Robust principal component
  analysis?'' \emph{J. ACM}, vol.~58, no.~3, pp. 1--37, 2011.

\bibitem{CandesRecht2009}
E.~J. Cand\`{e}s and B.~Recht, ``Exact matrix completion via convex
  optimization,'' \emph{Foundations of Computational Mathematics}, vol.~9,
  no.~6, pp. 717--772, 2009.

\bibitem{Fan2017290}
J.~Fan and T.~W. Chow, ``Matrix completion by least-square, low-rank, and
  sparse self-representations,'' \emph{Pattern Recognition}, vol.~71, pp. 290
  -- 305, 2017.

\bibitem{fan2019factor}
J.~Fan, L.~Ding, Y.~Chen, and M.~Udell, ``Factor group-sparse regularization
  for efficient low-rank matrix recovery,'' in \emph{Advances in Neural
  Information Processing Systems}, 2019, pp. 5104--5114.

\bibitem{cs_Donoho}
D.~L. {Donoho}, ``Compressed sensing,'' \emph{IEEE Transactions on Information
  Theory}, vol.~52, no.~4, pp. 1289--1306, April 2006.

\bibitem{SSC_PAMIN_2013}
E.~Elhamifar and R.~Vidal, ``Sparse subspace clustering: Algorithm, theory, and
  applications,'' \emph{IEEE Transactions on Pattern Analysis and Machine
  Intelligence}, vol.~35, no.~11, pp. 2765--2781, 2013.

\bibitem{LRR_PAMI_2013}
G.~Liu, Z.~Lin, S.~Yan, J.~Sun, Y.~Yu, and Y.~Ma, ``Robust recovery of subspace
  structures by low-rank representation,'' \emph{IEEE Transactions on Pattern
  Analysis and Machine Intelligence}, vol.~35, no.~1, pp. 171--184, 2013.

\bibitem{FAN201839}
J.~Fan, Z.~Tian, M.~Zhao, and T.~W. Chow, ``Accelerated low-rank representation
  for subspace clustering and semi-supervised classification on large-scale
  data,'' \emph{Neural Networks}, vol. 100, pp. 39 -- 48, 2018.

\bibitem{you2018scalable}
C.~You, C.~Li, D.~P. Robinson, and R.~Vidal, ``Scalable exemplar-based subspace
  clustering on class-imbalanced data,'' in \emph{Proceedings of the European
  Conference on Computer Vision (ECCV)}, 2018, pp. 67--83.

\bibitem{mairal2009online}
J.~Mairal, F.~Bach, J.~Ponce, and G.~Sapiro, ``Online dictionary learning for
  sparse coding,'' in \emph{Proceedings of the 26th annual international
  conference on machine learning}.\hskip 1em plus 0.5em minus 0.4em\relax ACM,
  2009, pp. 689--696.

\bibitem{mairal2009supervised}
J.~Mairal, J.~Ponce, G.~Sapiro, A.~Zisserman, and F.~R. Bach, ``Supervised
  dictionary learning,'' in \emph{Advances in neural information processing
  systems}, 2009, pp. 1033--1040.

\bibitem{zhang2019joint}
Z.~Zhang, J.~Ren, W.~Jiang, Z.~Zhang, R.~Hong, S.~Yan, and M.~Wang, ``Joint
  subspace recovery and enhanced locality driven robust flexible discriminative
  dictionary learning,'' \emph{IEEE Transactions on Circuits and Systems for
  Video Technology}, 2019.

\bibitem{zhang2019scalable}
Z.~Zhang, W.~Jiang, Z.~Zhang, S.~Li, G.~Liu, and J.~Qin, ``Scalable
  block-diagonal locality-constrained projective dictionary learning,'' in
  \emph{Proceedings of the 28th International Joint Conference on Artificial
  Intelligence}.\hskip 1em plus 0.5em minus 0.4em\relax AAAI Press, 2019, pp.
  4376--4382.

\bibitem{4483511}
J.~{Wright}, A.~Y. {Yang}, A.~{Ganesh}, S.~S. {Sastry}, and Y.~{Ma}, ``Robust
  face recognition via sparse representation,'' \emph{IEEE Transactions on
  Pattern Analysis and Machine Intelligence}, vol.~31, no.~2, pp. 210--227, Feb
  2009.

\bibitem{zhang2011sparse}
L.~Zhang, M.~Yang, and X.~Feng, ``Sparse representation or collaborative
  representation: Which helps face recognition?'' in \emph{Computer vision
  (ICCV), 2011 IEEE international conference on}.\hskip 1em plus 0.5em minus
  0.4em\relax IEEE, 2011, pp. 471--478.

\bibitem{lu2013online}
C.~Lu, J.~Shi, and J.~Jia, ``Online robust dictionary learning,'' in
  \emph{Proceedings of the IEEE Conference on Computer Vision and Pattern
  Recognition}, 2013, pp. 415--422.

\bibitem{jiang2015robust}
W.~Jiang, F.~Nie, and H.~Huang, ``Robust dictionary learning with capped
  {$\ell_1$}-norm,'' in \emph{Twenty-Fourth International Joint Conference on
  Artificial Intelligence}, 2015.

\bibitem{li2016weakly}
Z.~Li and J.~Tang, ``Weakly supervised deep matrix factorization for social
  image understanding,'' \emph{IEEE Transactions on Image Processing}, vol.~26,
  no.~1, pp. 276--288, 2016.

\bibitem{li2017robust}
Z.~Li, J.~Tang, and X.~He, ``Robust structured nonnegative matrix factorization
  for image representation,'' \emph{IEEE transactions on neural networks and
  learning systems}, vol.~29, no.~5, pp. 1947--1960, 2017.

\bibitem{LSSLRSC}
V.~M. Patel, N.~Hien~Van, and R.~Vidal, ``Latent space sparse and low-rank
  subspace clustering,'' \emph{Selected Topics in Signal Processing, IEEE
  Journal of}, vol.~9, no.~4, pp. 691--701, 2015.

\bibitem{Mika99kernelpca}
S.~Mika, B.~Schölkopf, A.~Smola, K.-R. Müller, M.~Scholz, and G.~Rätsch,
  ``Kernel pca and de-noising in feature spaces,'' in \emph{Advances in Neural
  Information Processing Systems 11}.\hskip 1em plus 0.5em minus 0.4em\relax
  MIT Press, 1999, pp. 536--542.

\bibitem{kwok2004pre}
J.-Y. Kwok and I.-H. Tsang, ``The pre-image problem in kernel methods,''
  \emph{IEEE transactions on neural networks}, vol.~15, no.~6, pp. 1517--1525,
  2004.

\bibitem{huang2009robust}
S.-Y. Huang, Y.-R. Yeh, and S.~Eguchi, ``Robust kernel principal component
  analysis,'' \emph{Neural computation}, vol.~21, no.~11, pp. 3179--3213, 2009.

\bibitem{nguyen2009robust}
M.~H. Nguyen and F.~Torre, ``Robust kernel principal component analysis,'' in
  \emph{Advances in Neural Information Processing Systems}, 2009, pp.
  1185--1192.

\bibitem{fan2019exactly}
J.~Fan and T.~W. Chow, ``Exactly robust kernel principal component analysis,''
  \emph{IEEE transactions on neural networks and learning systems}, 2019.

\bibitem{vincent2010stacked}
P.~Vincent, H.~Larochelle, I.~Lajoie, Y.~Bengio, and P.-A. Manzagol, ``Stacked
  denoising autoencoders: Learning useful representations in a deep network
  with a local denoising criterion,'' \emph{Journal of machine learning
  research}, vol.~11, no. Dec, pp. 3371--3408, 2010.

\bibitem{burger2012image}
H.~C. Burger, C.~J. Schuler, and S.~Harmeling, ``Image denoising: Can plain
  neural networks compete with bm3d?'' in \emph{2012 IEEE conference on
  computer vision and pattern recognition}.\hskip 1em plus 0.5em minus
  0.4em\relax IEEE, 2012, pp. 2392--2399.

\bibitem{mao2016image}
X.~Mao, C.~Shen, and Y.-B. Yang, ``Image restoration using very deep
  convolutional encoder-decoder networks with symmetric skip connections,'' in
  \emph{Advances in neural information processing systems}, 2016, pp.
  2802--2810.

\bibitem{gao2017demand}
R.~Gao and K.~Grauman, ``On-demand learning for deep image restoration,'' in
  \emph{Proceedings of the IEEE International Conference on Computer Vision},
  2017, pp. 1086--1095.

\bibitem{van2013design}
H.~Van~Nguyen, V.~M. Patel, N.~M. Nasrabadi, and R.~Chellappa, ``Design of
  non-linear kernel dictionaries for object recognition,'' \emph{IEEE
  Transactions on Image Processing}, vol.~22, no.~12, pp. 5123--5135, 2013.

\bibitem{7299018}
M.~{Harandi} and M.~{Salzmann}, ``Riemannian coding and dictionary learning:
  Kernels to the rescue,'' in \emph{2015 IEEE Conference on Computer Vision and
  Pattern Recognition (CVPR)}, June 2015, pp. 3926--3935.

\bibitem{liu2015robust}
H.~Liu, J.~Qin, H.~Cheng, and F.~Sun, ``Robust kernel dictionary learning using
  a whole sequence convergent algorithm,'' in \emph{Twenty-Fourth International
  Joint Conference on Artificial Intelligence}, 2015.

\bibitem{quan2016equiangular}
Y.~Quan, C.~Bao, and H.~Ji, ``Equiangular kernel dictionary learning with
  applications to dynamic texture analysis,'' in \emph{Proceedings of the IEEE
  Conference on Computer Vision and Pattern Recognition}, 2016, pp. 308--316.

\bibitem{aharon2006k}
M.~Aharon, M.~Elad, and A.~Bruckstein, ``K-{SVD}: An algorithm for designing
  overcomplete dictionaries for sparse representation,'' \emph{IEEE
  Transactions on signal processing}, vol.~54, no.~11, pp. 4311--4322, 2006.

\bibitem{skretting2010recursive}
K.~Skretting and K.~Engan, ``Recursive least squares dictionary learning
  algorithm,'' \emph{IEEE Transactions on Signal Processing}, vol.~58, no.~4,
  pp. 2121--2130, 2010.

\bibitem{mairal2011task}
J.~Mairal, F.~Bach, and J.~Ponce, ``Task-driven dictionary learning,''
  \emph{IEEE transactions on pattern analysis and machine intelligence},
  vol.~34, no.~4, pp. 791--804, 2011.

\bibitem{6516503}
Z.~{Jiang}, Z.~{Lin}, and L.~S. {Davis}, ``Label consistent {K-SVD}: Learning a
  discriminative dictionary for recognition,'' \emph{IEEE Transactions on
  Pattern Analysis and Machine Intelligence}, vol.~35, no.~11, pp. 2651--2664,
  Nov 2013.

\bibitem{chen2013robust}
Z.~Chen and Y.~Wu, ``Robust dictionary learning by error source
  decomposition,'' in \emph{Proceedings of the IEEE International Conference on
  Computer Vision}, 2013, pp. 2216--2223.

\bibitem{wang2013semi}
H.~Wang, F.~Nie, W.~Cai, and H.~Huang, ``Semi-supervised robust dictionary
  learning via efficient l-norms minimization,'' in \emph{Proceedings of the
  IEEE International Conference on Computer Vision}, 2013, pp. 1145--1152.

\bibitem{rdlhskf}
S.~P. Awate and N.~N. Koushik, ``Robust dictionary learning on the hilbert
  sphere in kernel feature space,'' in \emph{Machine Learning and Knowledge
  Discovery in Databases}, P.~Frasconi, N.~Landwehr, G.~Manco, and J.~Vreeken,
  Eds.\hskip 1em plus 0.5em minus 0.4em\relax Cham: Springer International
  Publishing, 2016, pp. 731--748.

\bibitem{8031891}
T.~{Zhou}, F.~{Liu}, H.~{Bhaskar}, and J.~{Yang}, ``Robust visual tracking via
  online discriminative and low-rank dictionary learning,'' \emph{IEEE
  Transactions on Cybernetics}, vol.~48, no.~9, pp. 2643--2655, Sep. 2018.

\bibitem{KSSC}
V.~M. {Patel} and R.~{Vidal}, ``Kernel sparse subspace clustering,'' in
  \emph{2014 IEEE International Conference on Image Processing (ICIP)}, Oct
  2014, pp. 2849--2853.

\bibitem{NIPS2013_4865}
Y.-X. Wang, H.~Xu, and C.~Leng, ``Provable subspace clustering: When lrr meets
  ssc,'' in \emph{Advances in Neural Information Processing Systems 26},
  C.~J.~C. Burges, L.~Bottou, M.~Welling, Z.~Ghahramani, and K.~Q. Weinberger,
  Eds.\hskip 1em plus 0.5em minus 0.4em\relax Curran Associates, Inc., 2013,
  pp. 64--72.

\bibitem{pmlr-v51-shen16}
J.~Shen and P.~Li, ``Learning structured low-rank representation via matrix
  factorization,'' in \emph{Proceedings of the 19th International Conference on
  Artificial Intelligence and Statistics}, ser. Proceedings of Machine Learning
  Research, vol.~51.\hskip 1em plus 0.5em minus 0.4em\relax Cadiz, Spain: PMLR,
  09--11 May 2016, pp. 500--509.

\bibitem{8573150}
M.~{Brbić} and I.~{Kopriva}, ``$\ell_0$ -motivated low-rank sparse subspace
  clustering,'' \emph{IEEE Transactions on Cybernetics}, vol.~50, no.~4, pp.
  1711--1725, April 2020.

\bibitem{fan2019online}
J.~Fan and M.~Udell, ``Online high rank matrix completion,'' in
  \emph{Proceedings of the IEEE Conference on Computer Vision and Pattern
  Recognition}, 2019, pp. 8690--8698.

\bibitem{CaiCandesShen2010}
J.-F. Cai, E.~J. Cand\`{e}s, and Z.~Shen, ``A singular value thresholding
  algorithm for matrix completion,'' \emph{SIAM Journal on Optimization},
  vol.~20, no.~4, pp. 1956--1982, 2010.

\bibitem{parikh2014proximal}
N.~Parikh, S.~Boyd \emph{et~al.}, ``Proximal algorithms,'' \emph{Foundations
  and Trends{\textregistered} in Optimization}, vol.~1, no.~3, pp. 127--239,
  2014.

\bibitem{bolte2014proximal}
J.~Bolte, S.~Sabach, and M.~Teboulle, ``Proximal alternating linearized
  minimization for nonconvex and nonsmooth problems,'' \emph{Mathematical
  Programming}, vol. 146, no. 1-2, pp. 459--494, 2014.

\bibitem{LSRSCLu2012}
C.-Y. Lu, H.~Min, Z.-Q. Zhao, L.~Zhu, D.-S. Huang, and S.~Yan, ``Robust and
  efficient subspace segmentation via least squares regression,'' in
  \emph{European conference on computer vision}.\hskip 1em plus 0.5em minus
  0.4em\relax Springer, 2012, pp. 347--360.

\bibitem{cai2014large}
D.~Cai and X.~Chen, ``Large scale spectral clustering via landmark-based sparse
  representation,'' \emph{IEEE transactions on cybernetics}, vol.~45, no.~8,
  pp. 1669--1680, 2014.

\bibitem{coil20}
S.~A. Nene, S.~K. Nayar, H.~Murase \emph{et~al.}, ``Columbia object image
  library (coil-20),'' 1996.

\bibitem{coil100}
S.~A. Nene, S.~K. Nayar, and H.~Murase, ``Columbia object image library
  (coil-100),'' 1996.

\bibitem{Dataset_ExtendYaleB}
L.~Kuang-Chih, J.~Ho, and D.~J. Kriegman, ``Acquiring linear subspaces for face
  recognition under variable lighting,'' \emph{IEEE Transactions on Pattern
  Analysis and Machine Intelligence}, vol.~27, no.~5, pp. 684--698, 2005.

\bibitem{ARfacedata}
A.~M. Mart{\'\i}nez and A.~C. Kak, ``{PCA} versus {LDA},'' \emph{IEEE
  transactions on pattern analysis and machine intelligence}, vol.~23, no.~2,
  pp. 228--233, 2001.

\bibitem{lipor2017subspace}
J.~Lipor, D.~Hong, Y.~S. Tan, and L.~Balzano, ``Subspace clustering using
  ensembles of $ k $-subspaces,'' \emph{arXiv preprint arXiv:1709.04744}, 2017.

\end{thebibliography}

\vspace{-10pt}
\begin{IEEEbiographynophoto}{Jicong Fan}
received his B.E and M.E degrees in Automation and Control Science \&
Engineering, from Beijing University of Chemical Technology, Beijing, P.R., China, in 2010 and
2013, respectively. From 2013 to 2015, he was a research assistant at the University of Hong
Kong. He received his Ph.D. degree in Electronic Engineering, from City University of Hong
Kong, Hong Kong S.A.R. in 2018. From 2018.01 to 2018.06, he was a visiting scholar at the
Department of Electrical and Computer Engineering, University of Wisconsin-Madison, USA. From 2018.10 to 2020.07, he was a Postdoc Associate at the School of Operations Research and Information
Engineering, Cornell University, Ithaca, USA. Currently, he is a Research Assistant Professor at the School of Data Science, The Chinese University of Hong Kong (Shenzhen) and Shenzhen Research Institute of Big Data, Shenzhen, China. His research interests include statistical process
control, signal processing, computer vision, optimization, and machine learning.
\end{IEEEbiographynophoto}

\vspace{-10pt}
\begin{IEEEbiographynophoto}{Chengrun Yang}
received his BS degree in Physics from Fudan University, Shanghai, China in 2016. Currently, he is a PhD student at the School of Electrical and Computer Engineering, Cornell University. His research interests include the application of low dimensional structures and active learning in resource-constrained learning problems.
\end{IEEEbiographynophoto}

\vspace{-10pt}
\begin{IEEEbiographynophoto}{Madeleine Udell}
Madeleine Udell is Assistant Professor of Operations Research and Information Engineering
and Richard and Sybil Smith Sesquicentennial Fellow at Cornell University.
She studies optimization and machine learning for large scale data analysis and control,
with applications in marketing, demographic modeling, medical informatics, engineering system design,
and automated machine learning.
Her work has been recognized by an NSF CAREER award, an Office of Naval Research (ONR) Young Investigator Award,
and an INFORMS Optimization Society Best Student Paper Award (as advisor).
Madeleine completed her PhD at Stanford University in
Computational $\&$ Mathematical Engineering in 2015
under the supervision of Stephen Boyd,
and a one year postdoctoral fellowship at Caltech
in the Center for the Mathematics of Information
hosted by Professor Joel Tropp.
\end{IEEEbiographynophoto}

\end{document}